%% file: main.tex
\documentclass[10pt,twocolumn,letterpaper]{article}

\usepackage{cvpr}              

\usepackage{tikz}

\definecolor{myred}{RGB}{192,0,0}


\definecolor{cvprblue}{rgb}{0.21,0.49,0.74}
\usepackage[pagebackref,breaklinks,colorlinks,allcolors=cvprblue]{hyperref}

\def\paperID{*****} 
\def\confName{CVPR}
\def\confYear{2025}

\title{ VL-Uncertainty: Detecting Hallucination in Large Vision-Language Model \\ via Uncertainty Estimation }

\author{
    Ruiyang Zhang ${^{1}}$ \quad Hu Zhang${^{2}}$ \quad Zhedong Zheng${^1}$\thanks{Correspondence to zhedongzheng@um.edu.mo.}
    \\
     $^1$ FST and ICI, University of Macau, China \\ $^2$ CSIRO Data61, Australia \\
     \small{\url{https://vl-uncertainty.github.io/}}
}

\begin{document}
\maketitle
\begin{tikzpicture}[remember picture,overlay]
\node[anchor=north west,xshift=1cm,yshift=-2.95cm] at (current page.north west) 
{\includegraphics[width=1.5cm]{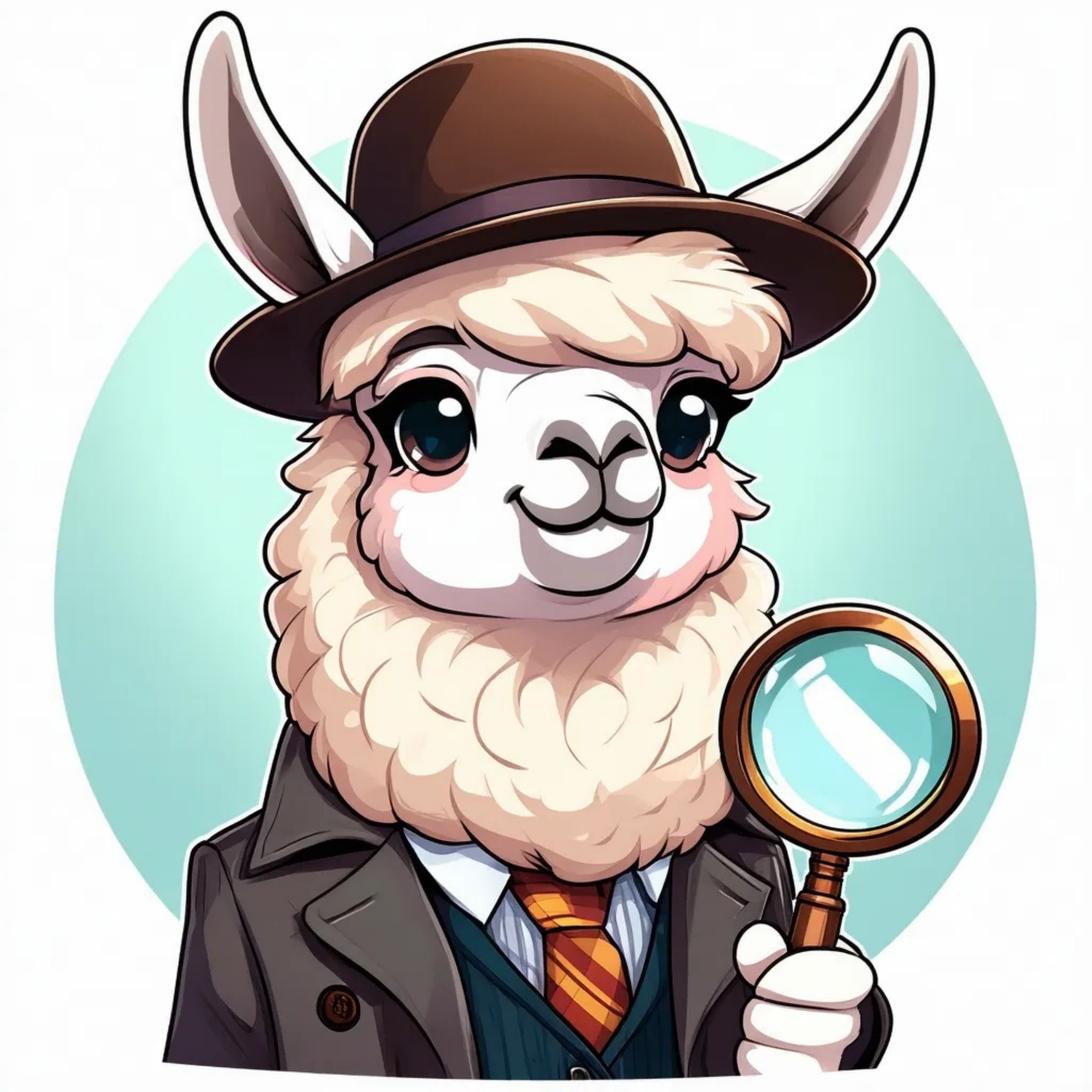}};
\end{tikzpicture}
\input{sec/0_abstract}    
\input{sec/1_intro}
\input{sec/2_related_work}
\input{sec/3_method}
\input{sec/4_exp}
\input{sec/5_conclusion}
\clearpage
{
    \small
    \bibliographystyle{ieeenat_fullname}
    \bibliography{main}
}
\input{sec/X_suppl}

\end{document}

%% file: sec/0_abstract.tex
\begin{abstract}

Given the higher information load processed by large vision-language models (LVLMs) compared to single-modal LLMs, detecting LVLM hallucinations requires more human and time expense, and thus rise a wider safety concerns. In this paper, we introduce VL-Uncertainty, the first uncertainty-based framework for detecting hallucinations in LVLMs. Different from most existing methods that require ground-truth or pseudo annotations, VL-Uncertainty utilizes uncertainty as an intrinsic metric. We measure uncertainty by analyzing the prediction variance across semantically equivalent but perturbed prompts, including visual and textual data. When LVLMs are highly confident, they provide consistent responses to semantically equivalent queries. However, when uncertain, the responses of the target LVLM become more random. Considering semantically similar answers with different wordings, we cluster LVLM responses based on their semantic content and then calculate the cluster distribution entropy as the uncertainty measure to detect hallucination. Our extensive experiments on 10 LVLMs across four benchmarks, covering both free-form and multi-choice tasks, show that VL-Uncertainty significantly outperforms strong baseline methods in hallucination detection.

\end{abstract}

%% file: sec/1_intro.tex
\section{Introduction}
\label{sec:intro}

Large vision-language models (LVLMs), capable of perceiving the world through diverse modalities, 
\eg, text, and images, have been widely applied in fields, \eg,  medical diagnosis~\cite{hu2024omnimedvqa,li2024llava,moor2023med}, embodied robotic~\cite{li2024manipllm,huang2023embodied,peng2023kosmos}, and autonomous driving~\cite{tian2024drivevlm,cui2024survey,wang2023drivemlm}. Despite their impressive performance, similar to large language models (LLMs)~\cite{huang2023survey,huang2023survey}, LVLMs inevitably generate hallucination with over confidence, if any, posing serious risks in safety-critical scenarios~\cite{bai2024hallucination,liu2024survey}. Compared to single-modal LLMs, detecting hallucination in LVLMs demands a deep understanding of multiple modalities~\cite{sahoo2024comprehensive}. It not only poses the challenges of the question understanding,  but also the difficulty in checking the answer authenticity. Therefore, researchers have resorted to the automatic hallucination detection.

\begin{figure}
    \centering
    \includegraphics[width=\linewidth]{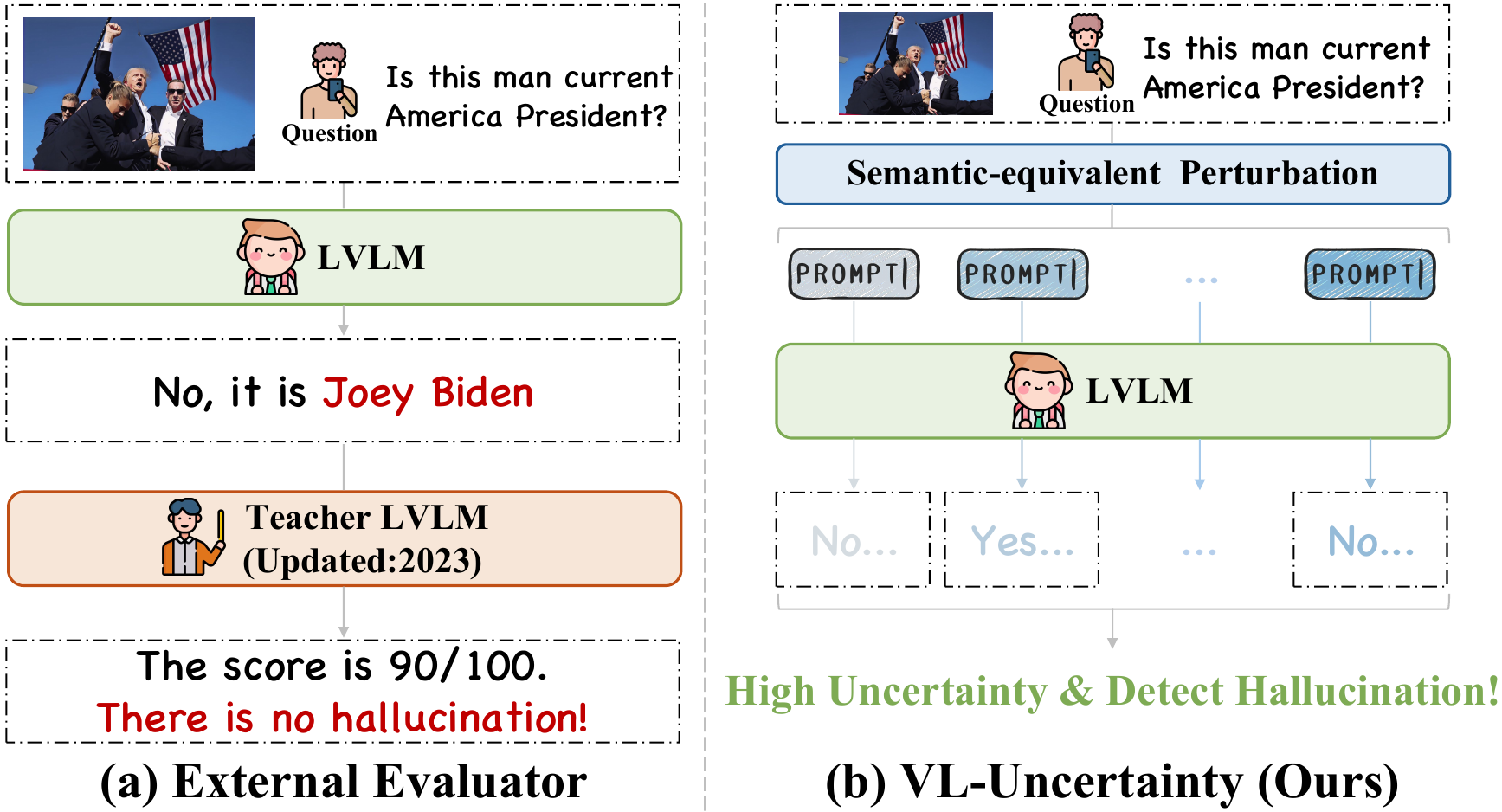}
    \caption{
    \textbf{Our motivation.}
    External evaluator-based methods usually suffer from knowledge missing when it comes to new domains (see (a)). In contrast, our VL-Uncertainty elicits intrinsic uncertainty of LVLM through proposed semantic-equivalent perturbation. Finally, refined uncertainty estimation facilitates reliable LVLM hallucination detection (see (b)).
    }
    \label{fig:motivation}
\end{figure}

Most existing works for LVLM hallucination detection are based on external knowledge sources~\cite{sun2023aligning,xiao2024detecting,liu2023mitigating}.
These methods can be coarsely divided into two families. One line of approaches~\cite{wang2024mfc,jing2023faithscore}
utilizes manually annotated ground truth, such as parsed real-world facts from knowledge databases~\cite{sun2023aligning}, to verify whether the responses of LVLMs are hallucinatory. 
Another line of methods relies on the pseudo annotations from extra models~\cite{liu2023mitigating,wang2023evaluation}. For instance, some works introduce
a `teacher-student' paradigm~\cite{xiao2024detecting}. The teacher LVLM takes the original question and the student answer as input and then scores the student answer. The student answers rating with low scores indicate hallucinations. In real-world scenarios, we, however, usually meet brand new problems, such as the impact of breaking news. We could apply the LVLM to give a prediction, but we do not know the probability of the hallucination. Both hallucination detection methods tend to fail, since we do not have the ground-truth reference, but we also can not rely on the out-of-the-date teacher LVLMs (see Fig.~\ref{fig:motivation}).

\begin{figure}
    \centering
    \includegraphics[width=\linewidth]{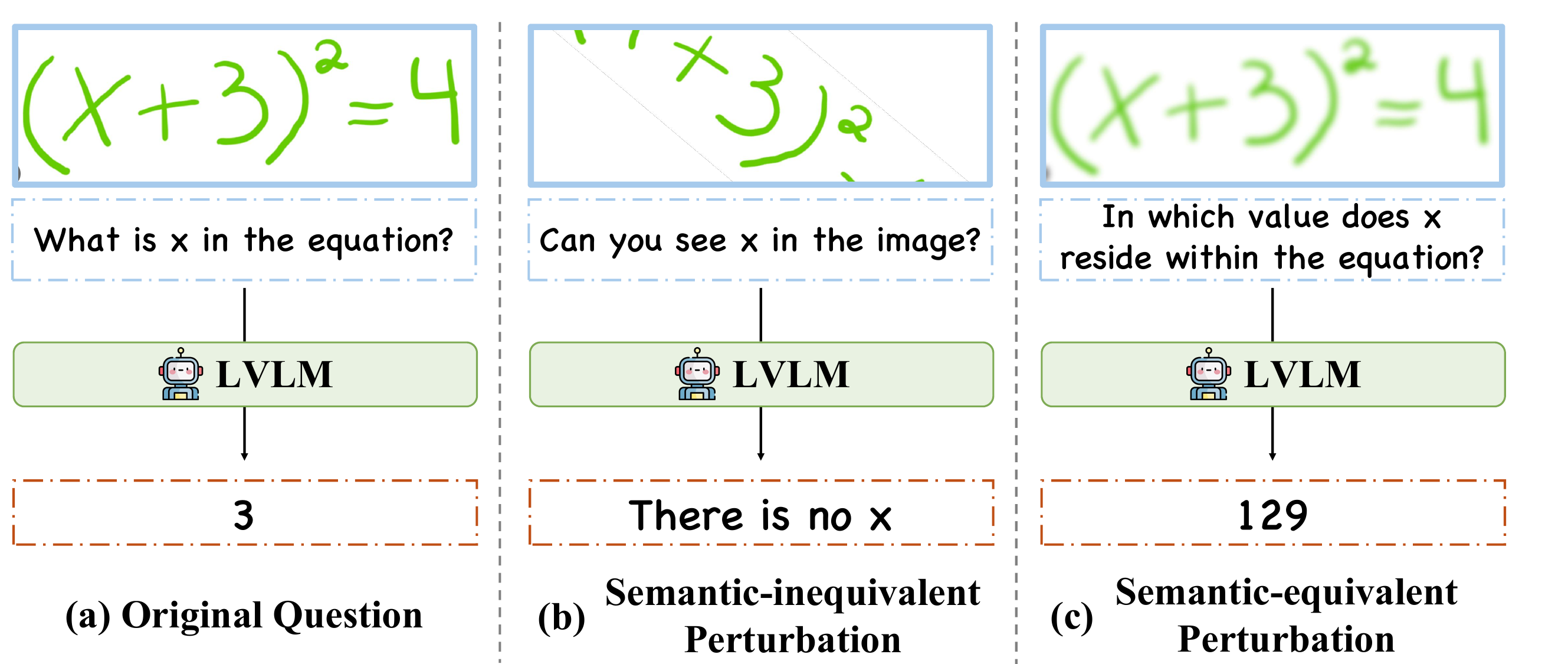}
    \vspace{-.2in}
    \caption{
    \textbf{Comparison between semantic-equivalent perturbations and inequivalent ones.}
    LVLMs inevitably generate hallucinatory answers (see (a)). While semantic-inequivalent perturbations yield correct answers, they do not provide insight into the uncertainty of LVLM for the original query, as shown in (b). In contrast, responses to semantically equivalent perturbed prompts, though potentially incorrect, offer valuable insight into the intrinsic uncertainty of LVLM. With only the exterior presentations of prompt altered, fluctuation of answers indicates elevated uncertainty (see (c)). This distinction highlights the utility of semantic-equivalent perturbations in assessing the reliability and consistency of LVLM responses.
    }
    \label{fig:equivalent}
    \vspace{-.1in}
\end{figure}

In an attempt to address this challenge, we propose VL-Uncertainty, the first uncertainty-based framework tailored for LVLM hallucination detection. Distinct from prior approaches that necessitate auxiliary information, our method intrinsically quantifies the uncertainty inherent to LVLM answers, enabling a mechanism for autonomous validation. Upon identifying elevated levels of uncertainty within an LVLM response, VL-Uncertainty categorizes the response as potentially hallucinatory. Specifically, we implement a technique involving semantic-equivalent perturbations to the prompts, thereby evaluating the uncertainty via the dispersion observed in the resulting answers. The foundational premise guiding this approach is that, under conditions of high confidence, LVLMs exhibit a tendency to generate consistent responses to queries that are semantically equivalent. Conversely, if perturbations that alter prompt exterior presentation lead to a divergence with responses of model, high uncertainty or potential hallucination is indicated (see Fig.~\ref{fig:equivalent}). In particular, we employ blurring as the semantic-inequivalent perturbation for visual prompts. Blurring maintains all elements of the original visual prompt and preserves underlying logic and meaning. This choice follows biological principles observed in the human visual system~\cite{baird2020myopia,haarman2020complications} and simulates the effect of varying distances between visual signals and the retina. For textual prompts, we deploy an off-the-shelf LLM to perturb the question without altering its meaning. By adjusting the temperature of the LLM, we control the degree of perturbation, analogous to visual blurring. 
These visual and textual prompts, paired by their levels of perturbation, are then fed into the LVLM to obtain a series of answers. Considering multiple LVLM answers with different wording, we first cluster the predicted answers by their semantics and calculate the entropy of the cluster distribution to quantify LVLM uncertainty as a continuous scalar. The uncertainty yielded by the series of answers enables the identification of varying levels of hallucination without extra models or manual annotations. We conduct experiments with 10 LVLMs across 4 benchmarks, encompassing both free-form and multi-choice formats. Our results show that VL-Uncertainty consistently surpasses strong baselines by clear margin in LVLM hallucination detection. 
Further qualitative analysis validates the superiority of VL-Uncertainty in effectively capturing LVLM uncertainty, thereby facilitating accurate hallucination detection.
In summary, our contributions are as follows:
\begin{itemize}
    \item We propose a new uncertainty-based framework, VL-Uncertainty, for detecting hallucination in Multi-modal Large Language Models (LVLMs). 
    We find that it is of importance to control the difficulty of prompts via semantic-equivalent perturbation, facilitating VL-Uncertainty capturing the randomness in LVLM response, indicating uncertainty and potential hallucination.
    Since it is an intrinsic metric, VL-Uncertainty could be easily scalable to new fields. 
    \item We conduct extensive experiments on 10 LVLMs across 4 benchmarks, including both free-form and multi-choice tasks. Our results show that VL-Uncertainty outperforms strong baselines in LVLM hallucination detection, thereby enhancing the safety and reliability of LVLM applications.
\end{itemize}

%% file: sec/2_related_work.tex
\section{Related Work}
\label{sec:related_work}

\noindent \textbf{Large Vision-Language Models.}
Early works primarily focus on generating text responses based on image and text prompts~\cite{liu2024visual,liu2024improved,zhu2023minigpt}. Building on these foundational efforts, subsequent studies have significantly extended the capabilities and application domains of LVLMs~\cite{wu2023next,yuan2024osprey,huang2023embodied}. Recent research has focused on refining prompt granularity from image-level to more detailed box- or point-level control~\cite{chen2023shikra,yuan2024osprey}. Based on these achievements, LVLMs have been applied in different fields, such as medical diagnosis~\cite{li2024llava,moor2023med}, embodied robotics~\cite{huang2023embodied,peng2023kosmos}, and autonomous driving~\cite{cui2024survey,wang2023drivemlm}. While these developments enhance LVLM capabilities, complex cross-modal interactions are introduced, compromising response reliability. In high-stakes applications, unreliable LVLM responses present significant safety risks, leading to high demands for accurate hallucination detection~\cite{hu2024omnimedvqa}. Distinct from existing approaches, we propose explicitly estimating intrinsic uncertainty of LVLM to facilitate hallucination detection, laying the foundation for safer human-LVLM interactions.

\begin{figure*}[!t]
    \vspace{-.2in}
    \centering
    \includegraphics[width=\textwidth]{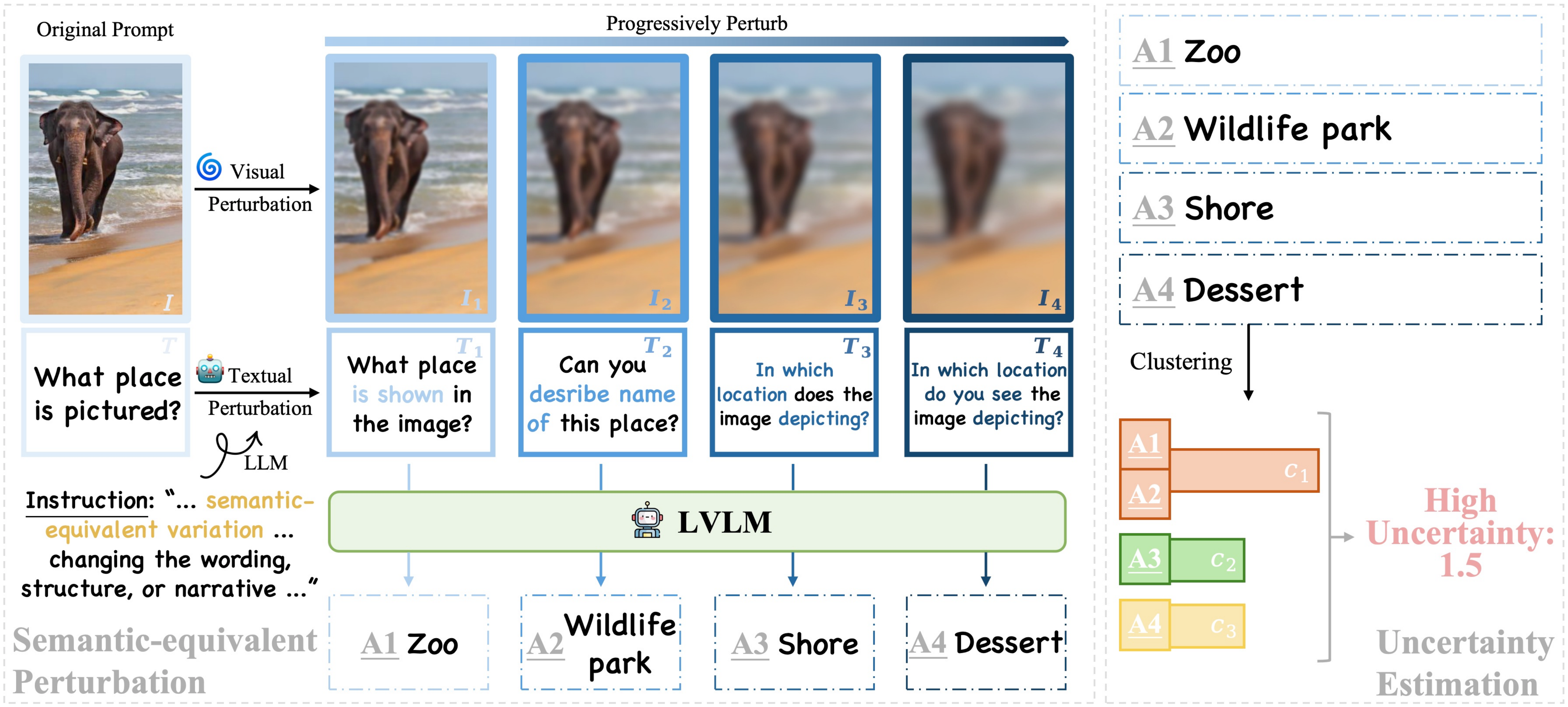}
    \caption{
    \textbf{Overall illustration of our proposed VL-Uncertainty.}
    To facilitate mining of uncertainty arising from various modalities, we apply semantic-equivalent perturbations \textbf{\emph{(left)}} to both visual and textual prompts. For visual prompt, the original image is blurred to varying degrees, mimicking human visual perception. For textual prompt, pre-trained LLM is prompted to rephrase the original question in semantic-equivalent manner with different temperatures. Detailed instruction is designed to achieve question rephrasing with the original semantic preserved. Prompt pairs with varying degrees of perturbation are harnessed to effectively elicit LVLM uncertainty. We cluster LVLM answer set by semantic meaning and utilize entropy of answer cluster distribution as LVLM uncertainty \textbf{\emph{(right)}}. The estimated uncertainty serves as a continuous indicator of different levels of LVLM hallucination.
    }
    \label{fig:method}
    \vspace{-.1in}
\end{figure*}

\noindent \textbf{Uncertainty Learning.}
Uncertainty learning methods~\cite{gawlikowski2023survey,he2023survey} generally fall into 
three primary categories. (1) Single deterministic methods~\cite{sensoy2018evidential,malinin2018predictive,zhang2024harnessing} modify a deterministic network to directly regress uncertainty. While these methods are straightforward and require minimal overhead, the predicted uncertainty, if no any regularization, has a potential to overfit all uncertain or very certain, compromising the training. (2) Bayesian methods~\cite{gal2016dropout,kendall2017uncertainties,welling2011bayesian,zheng2021rectifying,blundell2015weight} deploy stochastic Bayesian networks to quantify uncertainty by feeding the same input multiple times into one network with dynamic weights. Some works~\cite{gal2016dropout,gal2017concrete,miller2018dropout} leverages the dropout function, while others~\cite{yu2019robust, chen2024composed} explicitly introduce Gaussian noise. The variance between different predictions quantifies the uncertainty. The primary challenge lies in modeling Bayesian networks in a computationally efficient way. Following the spirit, test-time augmentation~\cite{lyzhov2020greedy,wang2019aleatoric,shanmugam2021better,kim2020learning,zhang2024ctrl} methods apply various augmentations to the input before feeding them to one single network. Similarly, variations in predictions due to these augmentations provide clues about uncertainty, though designing effective augmentations for meaningful uncertainty remains challenging. (3) Ensemble methods~\cite{lakshminarayanan2017simple,huang2017snapshot,malinin2019ensemble,wen2020batchensemble} conduct inference on multiple deterministic networks for the same input, with the entropy of the ensemble group predictions estimating uncertainty. However, memory and computational costs increase significantly with more ensemble members. Diverging from these existing methods, we propose estimating LVLM uncertainty based on semantic-equivalent perturbation on vision-language prompts and variance of corresponding answer set.

\noindent \textbf{LVLM Hallucination Detection.}
Research in this field can be divided into two main trajectories~\cite{liu2024survey,bai2024hallucination,li2024naturalbench}. (1) External-model-based evaluation: GAVIE~\cite{liu2023mitigating} leverages strong LVLM as a smart teacher to score responses of student LVLMs, with low scores indicating hallucinations. However, reliance on external pseudo annotations limits its application in unknown domains. More recently, HaELM~\cite{wang2023evaluation} specifically trains an LLM to score LVLM responses related to hallucinations. CCEval~\cite{zhai2023halle} suggests utilizing GPT4 API as intelligent parser to extract meaningful objects from responses and compare them with ground truth objects, although this introduces additional cost and resource. (2) Discrete rule-based checking: CHAIR~\cite{rohrbach2018object} suggests utilizing the discrete ratio of objects presented in the answer relative to a ground-truth object list to identify hallucinatory responses. However, this approach is restricted to the 80 COCO object classes. Building on CHAIR, POPE~\cite{li2023evaluating} optimizes the prompting technique by focusing on Yes-or-No questions, simplifying the checking process and improving evaluation stability. Unfortunately, treating hallucination detection purely as a binary classification task fails to capture the varying degrees of hallucinations.  Distinct from existing works, our proposed uncertainty estimation is entirely self-contained and free of external knowledge, thereby offering greater flexibility and robustness. Moreover, we explicitly estimate uncertainty within LVLMs to continuously indicate different levels of hallucination.

%% file: sec/3_method.tex
\section{Method}

\subsection{Semantic-equivalent Perturbation}
\label{sec:perturbation}

\noindent \textbf{Visual Prompts.}
Practically, we perturb the original input image multiple times by applying varying degrees of 2D Gaussian blurring (see Fig.~\ref{fig:method}). By adjusting blurring radius, we control the blur intensity from relatively clear to heavily blurred:
\begin{equation}
    I_{i} = \phi_{\text{vis}}(I, r_i),
\end{equation}
where $I$ denotes the original visual prompt, $r_i$ represents the radius of Gaussian blurring in the $i$-th perturbation, with $r_i < r_j$ for $i < j$. $\phi_{\text{vis}}$ refers to blurring operation and $I_i$ is the $i$-th perturbed visual prompt. $i$ ranges from 1 to $N$ and $N$ indicates the number of perturbations. Notably, blurring qualifies as a semantic-equivalent perturbation. It preserves the full content and structure of the original image, without introducing new objects or removing existing ones. This method maintains spatial information and the relationships between objects, given that it does not involve transformations like flipping or rotation. Visual attributes such as color, shape, and motion dynamics are also retained, ensuring the integrity of original image.

\noindent \textbf{Textual Prompts.}
For textual prompts, we also employ semantic-equivalent perturbations by varying the wording, grammatical structure, and narrative style without altering the underlying meaning (see Fig.~\ref{fig:method}). To achieve this, we utilize a pre-trained text-only LLM, prompting it to rephrase the original question while preserving its semantics. Specifically, we design detailed LLM instruction that focuses on varying words, structure, and narrative, while ensuring semantic equivalence. During each perturbation, we adjust the temperature of utilized LLM to achieve varying degrees of alteration, analogous to the visual perturbations:
\begin{equation}
    T_{i} = \phi_{\text{text}} \left( T,\tau_{i} \right),
\end{equation}
where $i = 1, 2, \ldots, N$, and $N$ is the number of perturbation times. $T$ represents the initial textual prompt, $\tau_{i}$ denotes the LLM temperature during the $i$-th perturbation, satisfying $\tau_i < \tau_j$ for $i < j$. $\phi_{\text{text}}$ refers to the utilized LLM and $T_{i}$ is the $i$-th perturbed textual prompt.

\noindent \textbf{Combination of Perturbed Prompts.}
We synchronize the perturbations of visual and textual prompts according to their respective degrees of perturbation. For visual prompts, the degree is quantified by the blurring radius, where a larger radius indicates heavier blurring. For textual prompts, the degree is determined by the LLM temperature, with higher temperatures leading to more narrative variations. Each visual prompt is perturbed several times, from low to high degrees, and similarly, textual prompts undergo comparable degrees of perturbation. Finally, we pair visual prompts with textual prompts that have been perturbed to a similar extent $\left\{ \langle I_{i}, T_{i} \rangle \mid i = 1, 2, \ldots, N \right\}$,
where $I_{i}$ and $T_{i}$ are the perturbed visual and textual prompts, respectively.

\noindent \textbf{Discussion.}
\textbf{Why is semantic-equivalent perturbation superior to semantic-inequivalent ones?}
Maintaining the original semantics of prompts during perturbation is essential. Semantically equivalent perturbation preserves the original meaning, ensuring that any fluctuations in responses stem directly from the inherent uncertainty of LVLM. This straightforward reflection of LVLM uncertainty enables more effective uncertainty estimation.
In contrast, if perturbations alter prompt semantics, responses could reasonably change, while those variations could not reflect true uncertainty. This can misleadingly increase entropy and falsely suggest high uncertainty. For example, even if the LVLM has low uncertainty about a question, altering the meaning of prompt can lead to varied responses, incorrectly indicating high uncertainty.
\textbf{What is the intuition behind utilizing image blurring?}
We select typical \textit{blurring} among various image perturbation techniques, drawing inspiration from human visual perception~\cite{baird2020myopia,haarman2020complications}. To illustrate this design, we take the nearsighted person as example. 
If their perception of an object remains stable regardless of whether they wear glasses, it indicates a low level of uncertainty about the object. Otherwise, it suggests a higher level of uncertainty about the object. 
Similarly, by applying varying degrees of blurring to visual prompts, we can measure the intrinsic uncertainty of LVLM: stable responses across different blur levels suggest low uncertainty, while significant changes in responses indicate higher uncertainty. Quantitative ablations further validate our intuition (see Table~\ref{tab:visual_perturbation}).

\subsection{Uncertainty Estimation}

We quantify LVLM uncertainty by measuring the variance within the set of generated answers (see Fig.~\ref{fig:method}). Notably, we consider the entropy across different semantics rather than mere lexical variations. Specifically, we use a pre-trained LLM to evaluate mutual semantic entailment between pairs of LVLM answers. A pair of answers is considered semantically entailed only if each answer entails the other. This operation is iteratively applied across the entire set of sampled answers, grouping them by underlying meaning. We thus obtain a set of semantic clusters $\{\boldsymbol{c_{i}}\}_{i=1}^{N_c}$, where $N_c$ is the total number of semantic clusters with $N_{c} \leq N$. Then, we calculate the entropy of the cluster distribution to estimate uncertainty:
\begin{equation}
    U_{\text{LVLM}} = - \sum_{i=1}^{N_{C}} p(\boldsymbol{c_{i}}) \log p(\boldsymbol{c_{i}}),
\end{equation}
where $\boldsymbol{c_{i}}$ is the $i$-th semantic cluster containing answers with same semantics, $p(\boldsymbol{c_{i}})$ denotes distribution probability of $i$-th semantic cluster. $U_{\text{LVLM}}$ represents the estimated LVLM uncertainty.

\noindent \textbf{Discussion.}
\textbf{Why estimate uncertainty on answer variances against prompt perturbations, rather than allowing LVLMs to directly regress uncertainty like previous works~\cite{xiong2023can}?}
We observe a severe over-confidence problem when allowing LVLMs to assign confidence scores to their own responses. For example, even with the prompt of hallucination, LVLM usually still regard their responses as absolutely correct and assign a high confidence score. This is similar to humans, \ie, individuals tend to overestimate their confidence without repeated consideration~\cite{magnus2018grade,klayman1999overconfidence}. To address this, we leverage prompt perturbations and multiple sampling to better capture LVLM uncertainty. By progressively perturbing prompts and sampling multiple times, we obtain a more refined uncertainty estimation.
\textbf{What are benefits of VL-Uncertainty over vanilla semantic entropy~\cite{farquhar2024detecting,kuhn2023semantic}?}
VL-Uncertainty is tailored for vision-language scenarios with semantically equivalent perturbations for each modality. The semantically equivalent perturbations enable fine-grained reflection of the LVLM on the given prompts. Additionally, the design of the image-text pairs is with increasing perturbations. Compared with the relatively random temperature design in~\cite{farquhar2024detecting,kuhn2023semantic}, the proposed method contains different levels of difficulty in the prompts of both modalities.

\subsection{LVLM Hallucination Detection}

Notably, our estimated uncertainty is continuous and proficient in indicating varying levels of hallucination, from minor deviations to complete logical incoherence. This continuous measure effectively captures the full spectrum of hallucinations encountered in LVLMs.
However, existing benchmarks lack such continuous, fine-grained ground truth. To obtain quantitative results and compare with previous methods, we establish a decision threshold for our estimated uncertainty: answers with uncertainty exceeding this threshold are predicted as hallucinatory, while those below it are considered not. Finally, we compare hallucination detection predictions with the hallucination ground truth labels to assess the accuracy of VL-Uncertainty in detecting hallucinations as $\left( N_{\text{F-N}} + N_{\text{T-P}}\right) / N_{\text{Total}}$, where $N_{\text{F-N}}$ (False-Negtive) refers to count of cases where answer is hallucinatory and VL-Uncertainty predict the answer as hallucinatory, $N_{\text{T-P}}$ (True-Positive) refers to count of cases where answer is non-hallucinatory and VL-Uncertainty predict the answer as non-hallucinatory. $N_{\text{Total}}$ is the total number of questions. This metric indicates the proportion of correct predictions (both hallucinatory and non-hallucinatory) made by VL-Uncertainty across all evaluated cases.

\noindent \textbf{Discussion.}
\textbf{Can VL-Uncertainty be applied to any LVLM in the image-text domain?}
VL-Uncertainty is a versatile and scalable hallucination detection framework that can be applied to any image-text LVLM. It leverages both the input prompts and output answers of LVLMs to enable effective uncertainty estimation and hallucination detection, regardless of the specific structure or design of the LVLMs. As a result, VL-Uncertainty offers greater flexibility and robustness.

%% file: sec/4_exp.tex
\begin{table*}[!t]
    \vspace{-.4in}
    \centering
    \small
    \resizebox{\textwidth}{!}{
    \begin{tabular}{c|c|c|c|c|c|c|c|c|c|c}
    \toprule
    \textbf{MM-Vet} & \textbf{Qwen2VL-2B} & \textbf{Qwen2VL-7B} & \textbf{Qwen2VL-72B} & \textbf{LLaVA1.5-7B} & \textbf{LLaVA1.5-13B} & \textbf{InternVL2-1B} & \textbf{InternVL2-8B} & \textbf{InternVL2-26B} & \textbf{LLaVANeXT-7B} & \textbf{LLaVANeXT-13B} \\
    \midrule
    GAVIE~\cite{liu2023mitigating} & 29.36 & 43.58 & 51.38 & 23.39 & 24.77 & 30.73 & 30.73 & 22.48 & 37.61 & 43.58 \\
    \midrule
    Semantic Entropy~\cite{farquhar2024detecting} & 60.55 & 57.80 & 62.84 & 72.48 & 79.36 & 72.94 & 55.05 & 58.72 & 61.01 & 72.48 \\
    \midrule
    VL-Uncertainty (ours) & \textbf{69.72} & \textbf{64.22} & \textbf{71.56} & \textbf{82.11} & \textbf{80.28} & \textbf{74.31} & \textbf{65.14} & \textbf{64.22} & \textbf{72.02} & \textbf{74.31} \\
    \midrule
    \textbf{LLaVA-Bench} & \textbf{Qwen2VL-2B} & \textbf{Qwen2VL-7B} & \textbf{Qwen2VL-72B} & \textbf{LLaVA1.5-7B} & \textbf{LLaVA1.5-13B} & \textbf{InternVL2-1B} & \textbf{InternVL2-8B} & \textbf{InternVL2-26B} & \textbf{LLaVANeXT-7B} & \textbf{LLaVANeXT-13B} \\    
    \midrule
    GAVIE~\cite{liu2023mitigating} & 25.00 & 26.67 & 40.00 & 15.00 & 20.00 & 30.00 & 31.67 & 31.67 & 45.00 & 35.00 \\
    \midrule
    Semantic Entropy~\cite{farquhar2024detecting} & 61.67 & 55.00 & 61.67 & 70.00 & 70.00 & 65.00 & 60.00 & 53.33 & 61.67 & 65.00 \\
    \midrule
    VL-Uncertainty (ours) & \textbf{70.00} & \textbf{68.33} & \textbf{71.67} & \textbf{83.33} & \textbf{78.33} & \textbf{71.67} & \textbf{66.67} & \textbf{73.33} & \textbf{68.33} & \textbf{68.33} \\
    \midrule
    \textbf{MMMU} & \textbf{Qwen2VL-2B} & \textbf{Qwen2VL-7B} & \textbf{Qwen2VL-72B} & \textbf{LLaVA1.5-7B} & \textbf{LLaVA1.5-13B} & \textbf{InternVL2-1B} & \textbf{InternVL2-8B} & \textbf{InternVL2-26B} & \textbf{LLaVANeXT-7B} & \textbf{LLaVANeXT-13B} \\    
    \midrule
    GAVIE~\cite{liu2023mitigating} & 37.82 & 48.36 & 57.09 & 37.58 & 44.61 & 40.61 & 48.12 & 33.21 & 43.64 & 45.82 \\
    \midrule
    Semantic Entropy~\cite{farquhar2024detecting} & 53.82 & 54.91 & 60.36 & 52.61 & 50.18 & 53.82 & 54.91 & 52.48 & 52.61 & 50.18 \\
    \midrule
    VL-Uncertainty (ours) & \textbf{58.91} & \textbf{59.76} & \textbf{65.58} & \textbf{56.05} & \textbf{53.62} & \textbf{56.36} & \textbf{55.15} & \textbf{58.91} & \textbf{59.27} & \textbf{54.90} \\
    \midrule
    \textbf{ScienceQA} & \textbf{Qwen2VL-2B} & \textbf{Qwen2VL-7B} & \textbf{Qwen2VL-72B} & \textbf{LLaVA1.5-7B} & \textbf{LLaVA1.5-13B} & \textbf{InternVL2-1B} & \textbf{InternVL2-8B} & \textbf{InternVL2-26B} & \textbf{LLaVANeXT-7B} & \textbf{LLaVANeXT-13B} \\
    \midrule
    GAVIE~\cite{liu2023mitigating} & 61.82 & 77.09 & 85.23 & 58.50 & 66.39 & 53.94 & 86.71 & 89.19 & 62.27 & 65.20 \\
    \midrule
    Semantic Entropy~\cite{farquhar2024detecting} & 54.04 & 77.94 & 87.06 & 61.77 & 68.02 & 64.45 & 90.08 & 91.32 & 67.67 & 65.34 \\
    \midrule
    VL-Uncertainty (ours) & \textbf{67.97} & \textbf{80.12} & \textbf{88.99} & \textbf{63.66} & \textbf{69.51} & \textbf{65.05} & \textbf{90.38} & \textbf{92.02} & \textbf{68.27} & \textbf{67.53} \\
    \bottomrule
    \end{tabular}
    }
    \vspace{-.1in}
    \caption{
    \textbf{Comparison with state-of-the-arts on both free-form benchmark (MM-Vet and LLaVABench) and multi-choice benchmark (MMMU and ScienceQA) for LVLM hallucination detection.}
    Our VL-Uncertainty yields significant improvements over strong baselines. This validates the efficacy of our proposed semantic-equivalent perturbation in eliciting and estimating LVLM uncertainty more accurately, which further facilitates LVLM hallucination detection. The reported results are hallucination detection accuracy. We re-implement semantic entropy~\cite{farquhar2024detecting} within vision-language context.
    }
    \vspace{-.1in}
    \label{tab:main}
\end{table*}

\section{Experiment}

\noindent \textbf{Benchmarks.}
Our experiments utilize both multi-choice and free-form benchmarks. For multi-choice benchmarks, we employ MMMU~\cite{yue2024mmmu} and  ScienceQA~\cite{lu2022learn}. MMMU presents a challenging set of college-level multi-modal questions spanning 30 subjects with 11.5K questions. ScienceQA comprises quiz questions typically found in American high school curricula, covering subjects like physics, chemistry, and biology, with a total of 21,208 samples split into training (12,726), validation (4,241), and testing set (4,241). For free-form benchmarks, we utilize MM-Vet~\cite{yu2023mm} and LLaVA-Bench~\cite{liu2024visual}, which include questions and answers in varied formats and lengths. MM-Vet, a recent benchmark, evaluates integrated LVLM capabilities across 6 basic abilities and 16 combinations, with 218 free-form question samples that span a range of topics. LLaVA-Bench, pioneering in assessing higher-level LVLM capabilities like logical reasoning, contains 60 distinct questions categorized into `convention', `detail', and `complexity'.

\noindent \textbf{LVLMs.}
We experiment with 10 LVLMs from 4 distinct model groups. Specifically, we utilize LLaVA1.5~\cite{liu2024visual}, LLaVA-NeXT~\cite{liu2024improved}, Qwen2VL~\cite{wang2024qwen2}, and InternVL2~\cite{chen2024internvl}. LLaVA1.5 introduces visual instruction tuning. It aligns a pre-trained vision encoder with an LLM through a projection layer and enables simultaneous processing of image and text. LLaVA-NeXT scales up the baseline model with richer data sources to enhance reasoning, video understanding, and world knowledge capabilities. Qwen2VL overcomes the limitations of predefined image resolution and enables LVLMs to handle various resolutions. InternVL2 focuses on scaling up the vision encoder within the alignment pipeline to improve general visual-language abilities.

\noindent \textbf{Implementation Details.}
We implement a unified codebase for LVLM uncertainty estimation and hallucination detection by including adopted benchmarks, LVLMs, baselines, and our VL-Uncertainty. Detailed settings for our VL-Uncertainty are as follows: (1) Initial answer generation. We set a low temperature of 0.1 for all LVLMs. The generated answer is compared with benchmark label to obtain whether this answer is hallucinatory. (2) Uncertainty estimation. We use a higher LVLM temperature to enable sampling process. In total, We perform 5 rounds of sampling. For visual perturbation, we employ 2D Gaussian blurring with radius in [0.6, 0.8, 1.0, 1.2, 1.4] to create different levels of image blur. For textual perturbation, we use a small LLM, Qwen2.5-3B-Instruct~\cite{bai2023qwen}, to rephrase questions, applying temperatures of [0.1, 0.2, 0.3, 0.4, 0.5]. The prompt for rephrasing is `Given the input question, generate a semantically equivalent variation by changing the wording, structure, grammar, or narrative. Ensure the perturbed question maintains the same meaning as the original.'. After we obtain sampled answer set, we use small LLM to check semantic entailment between answers. With answer set clustered by semantics, entropy of cluster distribution is calculated as uncertainty. (3) Hallucination detection. We utilize an uncertainty threshold of 1 for all experiments. If the estimated uncertainty is higher than the threshold, the initial answer is predicted by VL-Uncertainty as hallucination, while those lower are not. The hallucination predictions are compared with initial hallucination detection label (from (1)) to obtain hallucination detection accuracy. We utilize 2 H100 (80G) GPUs for all experiments. We also re-implement semantic-entropy~\cite{farquhar2024detecting} in the vision-language context since it is initially proposed in text-only domain.

\begin{table*}[t]
\vspace{-.4in}
\begin{subtable}{.33\linewidth}
\caption{} \label{tab:perturbation}
\centering
\resizebox{0.95\linewidth}{!}{
\begin{tabular}{c|c|c}
    \toprule
    \textbf{Visual Perturb.} & \textbf{Textual Perturb.} & \textbf{Hallu. Det. Acc.} \\
    \midrule
     & & 72.48 \\
    \midrule
    \checkmark & & 77.06 \\
    \midrule
     & \checkmark & 74.31 \\
    \midrule
    \checkmark & \checkmark & \textbf{82.11} \\
    \bottomrule
    \end{tabular}
}
\hfill
\centering 
    \caption{}
    \resizebox{0.95\linewidth}{!}{
    \begin{tabular}{c|c|c|c|c|c}
    \toprule
    \textbf{Visual Perturb.} & \textbf{Equiva.} & \textbf{Acc.} & \textbf{Visual Perturb.} & \textbf{Equiva.} & \textbf{Acc.} \\
    \midrule
    Rotation & $\times$ & 70.18 & GaussianNoise & \checkmark & 73.85 \\
    \midrule
    Flipping & $\times$ & 71.56 & Dropout & \checkmark & 73.85 \\
    \midrule
    Shifting & $\times$ & 72.02 & SaltAndPepper & \checkmark & 72.02 \\
    \midrule
    Cropping & $\times$ & 67.43 & Sharpen & \checkmark & 74.31 \\
    \midrule
    Erasing & $\times$ & 64.68 & AdjustContrast & \checkmark & 71.56 \\
    \midrule
    AdjustBrightness & \checkmark & 72.94 & Blurring & \checkmark & \textbf{82.11} \\
    \bottomrule
    \end{tabular}
    }
    \label{tab:visual_perturbation}
\end{subtable}   
\begin{subtable}{.33\linewidth}
\centering 
    \caption{}
    \resizebox{0.85\linewidth}{!}{
\begin{tabular}{c|c|c}
    \toprule
    \textbf{Textual Perturb.} & \textbf{Sem. Equiva.} & \textbf{Hallu. Det. Acc.} \\
    \midrule
    Swapping & $\times$ & 74.77 \\
    \midrule
    Deleting & $\times$ & 70.64 \\
    \midrule
    Inserting & $\times$ & 67.43 \\
    \midrule
    Replacing & $\times$ & 71.56 \\
    \midrule
    LLM Repharsing & \checkmark & \textbf{82.11} \\
    \bottomrule
    \end{tabular}
    }
    \label{tab:textual_perturbation}
\centering
\caption{}
\label{tab:blurring}\resizebox{0.85\linewidth}{!}{
\begin{tabular}{c|c|c}
    \toprule
    \textbf{Blurring Radius} & $\boldsymbol{\Delta}$ & \textbf{Hallu. Det. Acc.} \\
    \midrule
    $[0.1, 0.2, 0.3, 0.4, 0.5]$ & 0.1 & 75.69 \\
    \midrule
    $[0.6, 0.7, 0.8, 0.9, 1.0]$ & 0.1 & 74.31 \\
    \midrule
    $[0.6, 0.8, 1.0, 1.2, 1.4]$ & 0.2 & \textbf{82.11} \\
    \midrule
    $[0.5, 1.0, 1.5, 2.0, 2.5]$ & 0.5 & 76.15 \\
    \bottomrule
    \end{tabular}
}
\end{subtable}
\begin{subtable}{.33\linewidth}
\centering
\caption{}
\label{tab:llm_temperature}\resizebox{0.9\linewidth}{!}{
\begin{tabular}{c|c|c}
    \toprule
    \textbf{LLM Temperature} & $\boldsymbol{\Delta}$ & \textbf{Hallu. Det. Acc.} \\
    \midrule
    $[0.01, 0.02, 0.03, 0.04, 0.05]$ & 0.01 & 76.61 \\
    \midrule
    $[0.05, 0.1, 0.15, 0.2, 0.25]$ & 0.05 & 74.31 \\
    \midrule
    $[0.1, 0.2, 0.3, 0.4, 0.5]$ & 0.1 & \textbf{82.11} \\
    \midrule
    $[0.2, 0.4, 0.6, 0.8, 1.0]$ & 0.2 & 78.44 \\
    \midrule
    $[0.4, 0.8, 1.2, 1.6, 2.0]$ & 0.4 & 77.52 \\
    \bottomrule
    \end{tabular}
}
\hfill
\centering
\caption{}
\label{tab:llm}\resizebox{0.9\linewidth}{!}{
\begin{tabular}{c|c|c}
    \toprule
    \textbf{LLM Structure} & $\#$\textbf{Param.} & \textbf{Hallu. Det. Acc.} \\
    \midrule
    Qwen2.5-0.5B-Instruct & 0.5B & 50.92 \\
    \midrule
    Qwen2.5-1.5B-Instruct & 1.5B &69.27 \\
    \midrule
    Qwen2.5-3B-Instruct & 3B &\textbf{82.11} \\
    \midrule
    Qwen2.5-7B-Instruct & 7B &73.39 \\
    \bottomrule
    \end{tabular}
}
\end{subtable}
\vspace{-.1in}
\caption{
\textbf{Ablation studies on MM-Vet with LLaVA1.5-7B.}
\textbf{(a)} Ablation study of semantic-equivalent perturbation design. Perturbations applied across both modalities (visual and textual) yield the best results. Notably, the interaction between perturbed visual and textual prompts enables effective mining of uncertainty in complex vision-language context, thereby facilitating more refined LVLM hallucination detection.
\textbf{(b)} Ablation study of semantic-equivalent and semantic-inequivalent visual perturbation. Semantic-equivalent visual perturbation, such as blurring, proves superior to all other semantic-inequivalent perturbations. This underscores the importance of preserving the original semantics of visual prompts during perturbation, which more effectively elicits LVLM uncertainty.
\textbf{(c)} Ablation study of semantic-equivalent and semantic-inequivalent textual perturbation. Among all textual perturbations, LLM rephrasing yields optimal results. Other rule-based perturbations fail to maintain the original semantics of textual prompts, resulting in unsatisfactory outcomes.
\textbf{(d)} Ablation study of blurring radius in visual perturbation. Utilizing blurring radii of $[0.6, 0.8, 1.0, 1.2, 1.4]$ for visual perturbations yields the best results. Radius with medium gap, such as 0.2, ensures a reasonable variance between perturbed visual prompts, thereby more effectively eliciting LVLM uncertainty.
\textbf{(e)} Ablation of LLM temperature in textual perturbation. LLM temperatures of $[0.1, 0.2, 0.3, 0.4, 0.5]$ during perturbation yield the best results. This indicates that adjustments within a controlled range facilitate more effective elicitation of LVLM uncertainty.
\textbf{(f)} Ablation study of LLM for textual perturbation. Qwen2.5-3B-Instruct achieves best results among this LLM group.
}
\vspace{-.1in}
\end{table*}

\subsection{Comparison with State-of-the-arts}

We first present our hallucination detection results on the free-form question benchmarks (MM-Vet and LLaVABench) (see Table~\ref{tab:main}). Our VL-Uncertainty consistently achieves notable improvements over strong baselines~\cite{farquhar2024detecting,kuhn2023semantic} across various LVLM architectures and model sizes. Specifically, we observe +10.09\% for InternVL2-8B, +9.17\% for Qwen2VL-2B, and +6.42\% for Qwen2VL-7B on MM-Vet. These results validate the effectiveness of our proposed semantic-equivalent perturbation on both visual and textual prompts in enhancing LVLM uncertainty estimation and thereby facilitating hallucination detection.

We also present our hallucination detection results on multi-choice benchmarks (ScienceQA and MMMU) in Table~\ref{tab:main}. The consistent improvements over strong baselines validate the robustness of VL-Uncertainty across various benchmarks. On ScienceQA, VL-Uncertainty outperforms baselines by clear margins within Qwen2VL~\cite{wang2024qwen2} model group, achieving gains of +6.15\%, +2.18\%, and +1,93\% for 2B, 7B, and 72B models, respectively. Furthermore, VL-Uncertainty achieves a high hallucination detection accuracy of 92.02\% for InternVL2-26B, illustrating its substantial potential for effective hallucination detection.

\subsection{Ablation Studies and Further Discussion}

\noindent \textbf{Separated visual and textual semantic-equvialent perturbation.} We present the ablation results of our proposed semantic-equivalent perturbation in Table~\ref{tab:perturbation}. Single modality perturbation already yields substantial improvement compared to the vanilla baseline, with visual perturbation improving by +4.58\% and textual perturbation by +1.83\%. When semantic-equivalent perturbation is applied to both visual and textual prompts, VL-Uncertainty achieves optimal results with a performance of 82.11\%, surpassing a strong baseline by a clear margin (+9.63\%). This significant improvement validates the efficacy of our proposed perturbation approach in estimating LVLM uncertainty and enhancing LVLM hallucination detection.

\noindent \textbf{Semantic-equivalent and inequivalent visual perturbations.} We present a comparison between semantic-equivalent and semantic-inequivalent visual perturbations in Table~\ref{tab:visual_perturbation}. Specifically, we implement several baselines: `Rotation' rotates the original image with degrees in [-40, -20, 10, 20, 40]. `Flipping' utilizes 2 horizontal flipped and 3 vertical flipped images. `Shifting' moves the image up, down, left, and right within reasonable ranges. `Cropping' adopts crop ratios in [0.95, 0.9, 0.85, 0.8, 0.75] regarding the original size. `Erasing' randomly erases a square area with lengths in [50, 100, 150, 200, 250]. `GaussianNoise' adds per-channel noise with scale in [0.05, 0.1, 0.15, 0.2, 0.25]. `Dropout' randomly changes pixels to black with rate of [0.05, 0.1, 0.15, 0.2, 0.25]. `SaltAndPepper' is similar to `Dropout' but changes some pixels to white. `Sharpen' enhances the original image with a degree in [0.1, 0.2, 0.3, 0.4, 0.5]. `AdjustBrightness' and `AdjustContrast' alter the corresponding property by a factor of [0.8, 0.9, 1.1, 1.2, 1.3]. Notably, Semantic-equivalent visual perturbation, such as blurring, yields optimal results by a clear margin. This confirms the effectiveness of preserving original visual semantics during perturbation. Conversely, semantic-inequivalent techniques, with actual semantics of visual prompt altered, typically result in unsatisfactory outcomes.

\noindent \textbf{Semantic-equivalent and inequivalent textual perturbations.} We further report an ablation study on comparing semantic-equivalent and inequivalent textual perturbations (see Table.~\ref{tab:visual_perturbation}). For baselines: `Swapping' randomly swaps two words in the question; `Deleting' randomly deletes one word; `Inserting' inserts one word at a random place; and `Replacing' randomly changes one word with another. The observed pattern is similar to that in Table~\ref{tab:visual_perturbation}: semantic-equivalent perturbations surpass the inequivalent ones by a clear margin. This confirms retaining the original semantics of textual prompts contributes to accurate uncertainty estimation and effective hallucination detection.

\noindent \textbf{Design details of visual perturbation.} We also report an ablation study on design details for visual perturbation, focusing specifically on the blurring radius applied in different perturbations (see Table~\ref{tab:blurring}). We observe that maintaining a reasonable variance between perturbed prompts is crucial for achieving better performance. In visual perturbation, a blurring radius gap of 0.2 yields the best results. Conversely, both excessively small and overly large radius gaps negatively impact performance. A minimal gap fails to provide a sufficient difference between perturbed visual prompts, limiting the effective mining of uncertainty in the visual modality. On the other hand, a large gap introduces excessive variance, leading to inflated uncertainty that hinders accurate LVLM hallucination detection.

\begin{figure*}
    \centering
    \vspace{-.4in}
    \includegraphics[width=0.95\linewidth]{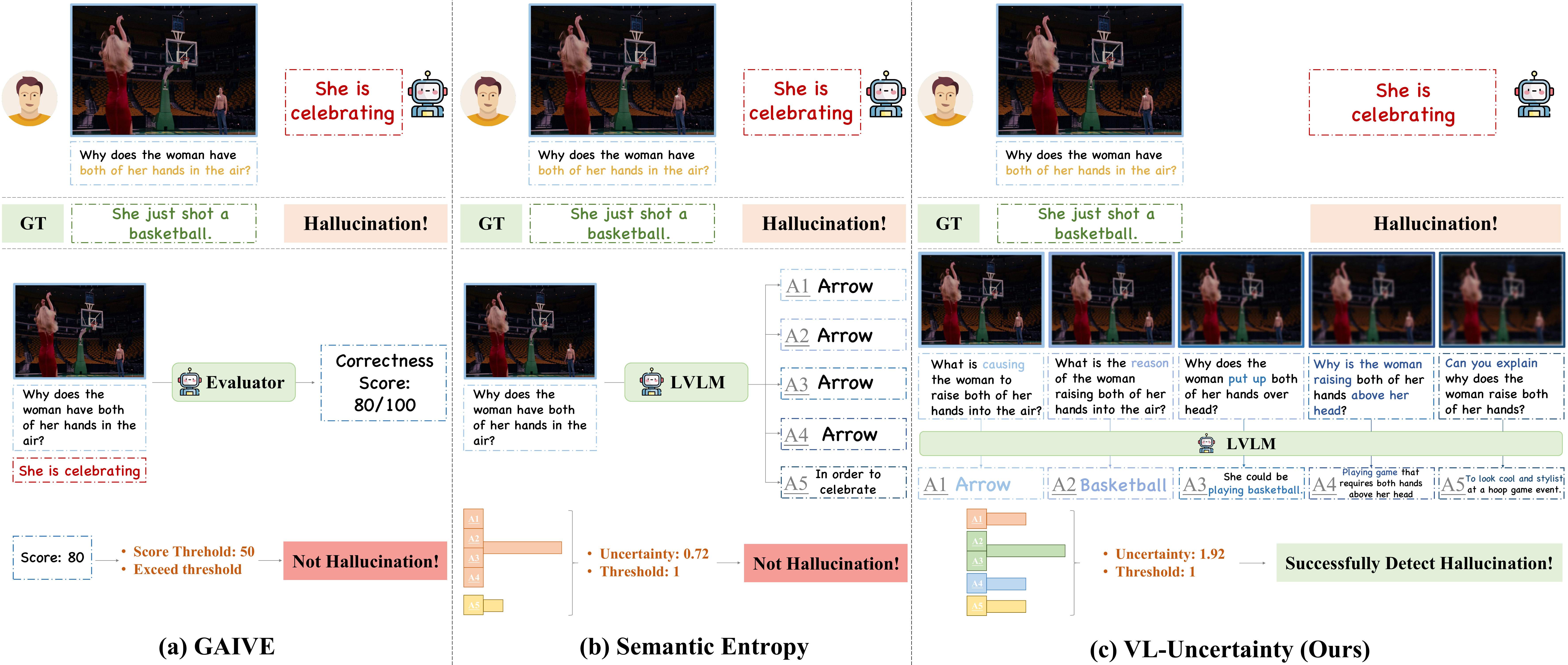}
    \vspace{-.1in}
    \caption{
    \textbf{Qualitative comparison between VL-Uncertainty and baselines.}
    We present a sample from free-form benchmark. For this hallucinatory sample, pseudo-annotation-based method~\cite{liu2023mitigating} fails to interpret the hidden-behind logic and thus misses detecting hallucination (see (a)). On the other hand, for semantic-entropy~\cite{farquhar2024detecting}, vanilla multi-sampling proves ineffective for mining LVLM uncertainty (see (b)). In contrast, our proposed semantic-equivalent perturbation on both visual and textual prompts successfully elicits LVLM uncertainty. This refined uncertainty estimation enhances the successful detection of LVLM hallucination (see (c)).
    }
    \label{fig:qualitative}
    \vspace{-.1in}
\end{figure*}

\noindent \textbf{Design details of textual perturbation.} We also present the ablation study on LLM temperature settings during textual perturbations in Table~\ref{tab:llm_temperature}. The best results from temperatures $[0.1, 0.2, 0.3, 0.4, 0.5]$ for different perturbations, maintaining a medium temperature gap of 0.1. This pattern is similar to that observed in visual perturbations (see Table~\ref{tab:blurring}), where a moderate gap yields the best results, both very small or large radius gaps compromise performance.

\noindent \textbf{Choices of LLMs for textual perturbation.} Additionally, we conduct the ablation study on LLMs in textual perturbation. We found that Qwen2.5-3B-Instruct is the optimal choice among models tested, as it balances the ability to generate semantically equivalent variations without introducing unrelated details. This model capacity is well-suited to maintaining semantic integrity in perturbed prompts, resulting in more accurate uncertainty estimation and improved hallucination detection. In contrast, smaller-capacity LLMs struggle to perform semantic-equivalent perturbation effectively on complex questions, degrading hallucination detection. Larger-capacity LLMs, on the other hand, tend to add unnecessary details to perturbed prompts, which hinders accurate uncertainty estimation.

\begin{figure}
    \centering
    \includegraphics[width=0.85\linewidth]{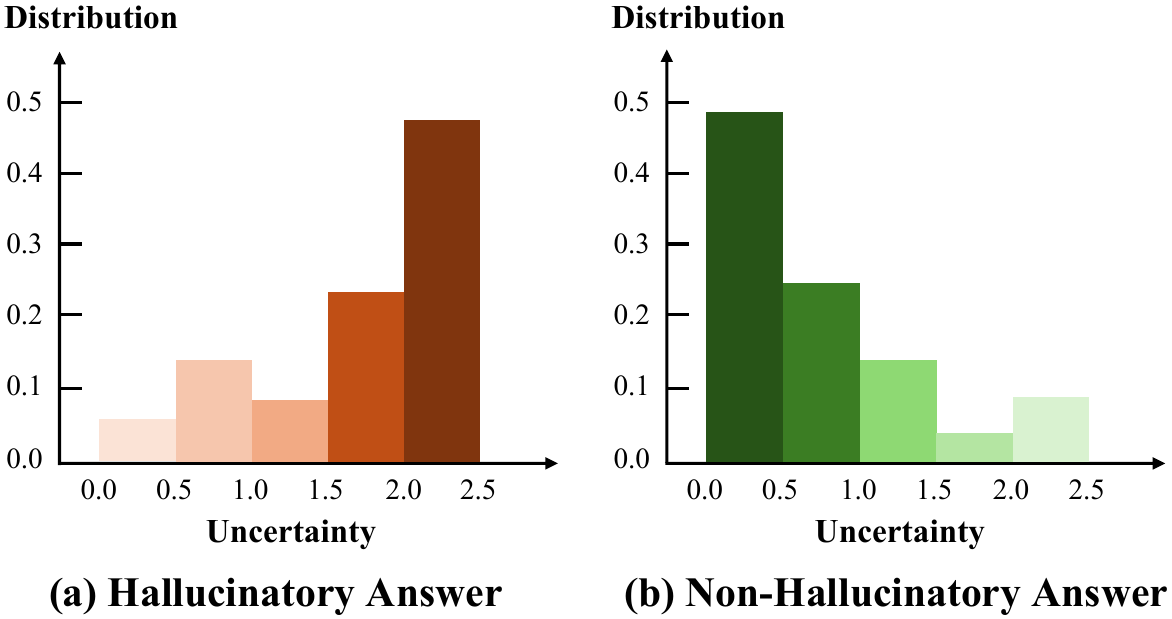}
    \vspace{-.1in}
    \caption{\textbf{Uncertainty distribution for hallucinatory and non-hallucinatory LVLM answers on MMVet.} Our VL-Uncertainty accurately assigns high uncertainty to hallucinatory answers and low uncertainty to non-hallucinatory answers. This distinct uncertainty distribution gap facilitates LVLM hallucination detection.
    }
    \label{fig:distribution}
    \vspace{-.15in}
\end{figure}

\subsection{Qualitative Analysis}
We present a qualitative analysis comparing VL-Uncertainty with the vanilla baselines~\cite{liu2023mitigating,farquhar2024detecting} in Fig.~\ref{fig:qualitative}. We present hallucinatory sample from free-form benchmarks. For GAVIE~\cite{liu2023mitigating}, external evaluator itself fails to interpret the underlying logic and thus misses detecting hallucination. In semantic entropy~\cite{farquhar2024detecting}, simply increasing the temperature during uncertainty estimation is insufficient to effectively capture LVLM uncertainty. The LVLM consistently produces similar answers for hallucination cases, leading to inaccurate uncertainty estimation and suboptimal hallucination detection (see (b)). In contrast, VL-Uncertainty, through our proposed semantic-equivalent perturbation, successfully captures high uncertainty and detects hallucinations in LVLMs (see (c)).

Fig.~\ref{fig:distribution} shows the statistical distribution of estimated uncertainty for hallucinatory and non-hallucinatory LVLM answers. Our estimated uncertainty closely calibrates with the accuracy of LVLM predictions: VL-Uncertainty predominantly assigns high uncertainty to hallucinatory answers, while assigning relatively low uncertainty to non-hallucinatory answers. The distinct gap in uncertainty distribution between hallucinatory and non-hallucinatory answers facilitates effective LVLM hallucination detection.

%% file: sec/5_conclusion.tex
\section{Conclusion}

In this paper, we introduce VL-Uncertainty, the first uncertainty-based framework for detecting LVLM hallucinations. Distinct from existing approaches based on external knowledge, VL-Uncertainty harnesses LVLM uncertainty as an intrinsic metric to identify hallucination. Recognizing the complexities inherent in multi-modal contexts, we propose semantic-equivalent perturbations for both visual and textual prompts. For visual prompts, we apply blurring at different levels, inspired by human visual processing. For textual prompts, a pre-trained LLM rephrases questions without altering their semantic meaning. Pairs of perturbed prompts with varying perturbations are utilized to effectively elicit LVLM uncertainty. Variance of corresponding answer semantics is harnessed to quantify LVLM uncertainty. Notably, our uncertainty serves as a continuous indicator proficient in illustrating varying levels of LVLM hallucinations. Through experiments with 10 LVLMs across 4 benchmarks (free-form and multi-choice), our VL-Uncertainty consistently demonstrates clear and substantial improvements over strong baselines, validating its effectiveness in LVLM hallucination detection.

%% file: sec/X_suppl.tex
\clearpage
    \setcounter{page}{1}
\maketitlesupplementary

\section{More Discussions}
\noindent \textbf{Why not conduct \textit{fine-tuning} to lower LVLM uncertainty and mitigate LVLM hallucination?}  
In this work, we focus on the post-processing of various LVLMs for hallucination detection, with fine-tuning being beyond the scope of our work. However, we believe the uncertainty estimated by VL-Uncertainty can benefit the fine-tuning process. The intrinsic uncertainty of LVLMs can serve as a valuable cue for adaptive weighting of different fine-tuning instructions, \emph{e.g.}, upsampling samples with high uncertainty to enhance learning efficacy. By incorporating uncertainty-aware fine-tuning or even pretraining, LVLMs could better focus on challenging cases, thereby mitigating potential hallucinations.

\noindent \textbf{Why not specifically consider hallucination mitigation in this work?}  
Our work primarily focuses on the accurate detection of hallucinations in LVLMs. While hallucination mitigation is a broad field encompassing various methods and techniques, we argue that it lies outside the scope of this study. Nevertheless, VL-Uncertainty, with its ability to accurately detect hallucinations in LVLMs, can lay a solid foundation for effective hallucination mitigation processes. Accurate hallucination detection can significantly facilitate follow-up tasks such as hallucination mitigation, reflection, and LVLM self-iteration. Furthermore, the scalability of VL-Uncertainty across diverse LVLM architectures can support the development of more general frameworks for hallucination mitigation.

\noindent \textbf{Why utilize progressively perturbed prompt pairs?}  
Our progressive perturbation strategy is detailed in Sec~\ref{sec:perturbation}. The final prompt pairs are denoted as $\left\{ \langle I_{i}, T_{i} \rangle \mid i = 1, 2, \ldots, N \right\}$, where $\langle I_{i}, T_{i} \rangle$ have lower perturbation degrees than $\langle I_{j}, T_{j} \rangle$ for $i < j$. We observe that progressive perturbation yields the best results (see Table.~\ref{tab:supp_perturbation_pairing}). In other pairing orders, such as random shuffling, the obtained prompts can consist of a hard prompt from one modality and a easy prompt from another modality. Those pairs could be equally challenging for the LVLM and result in all incorrect answers, leaving no variance in the answer set. As a result, uncertainty can not be effectively observed. In contrast, progressive perturbation ensures that both easy and hard prompt pairs are fed into the LVLM. For easy prompts, the LVLM generates correct answers, while for hard prompts, it generates incorrect answers. This fluctuation in responses makes high uncertainty evident.

\begin{table}[!t]
    \centering
    \resizebox{\linewidth}{!}{
    \begin{tabular}{c|c|c}
    \toprule
    \textbf{Visual Prompt Order} & \textbf{Textual Prompt Order} & \textbf{Hallucination Detection Accuraccy} \\
    \midrule
    $[I_1,I_2,I_3,I_4,I_5]$ & Random Shuffle & 73.85 \\
    \midrule
    $[I_1,I_2,I_3,I_4,I_5]$ & $[T_2,T_3,T_4,T_5,T_1]$ & 72.94 \\
    \midrule
    $[I_1,I_2,I_3,I_4,I_5]$ & $[T_3,T_4,T_5,T_1,T_2]$ & 75.69 \\
    \midrule
    $[I_1,I_2,I_3,I_4,I_5]$ & $[T_4,T_5,T_1,T_2,T_3]$ & 75.69 \\
    \midrule
    $[I_1,I_2,I_3,I_4,I_5]$ & $[T_5,T_4,T_3,T_2,T_1]$ & 74.77 \\
    \midrule
    $[I_1,I_2,I_3,I_4,I_5]$ & $[T_1,T_2,T_3,T_4,T_5]$ & \textbf{82.11} \\
    \bottomrule
    \end{tabular}
    }
    \caption{
    \textbf{Ablation of pairing order of visual and textual prompts with various degrees of perturbation.}  
    We observe that pairing visual and textual prompts in the same order yields the best results. This type of pairing produces prompt pairs with progressively perturbed prompts. This characteristic effectively challenges the LVLM and facilitates reliable uncertainty estimation.
    }
    \label{tab:supp_perturbation_pairing}
\end{table}

\begin{table}[!t]
    \centering
    \resizebox{\linewidth}{!}{
    \begin{tabular}{c|c|c}
    \toprule
    \textbf{Visual Perturbation} & \textbf{Semantic-equivalent} & \textbf{Hallucination Detection Accuracy} \\
    \midrule
    Rotate \& Shift & $\times$ & 73.85 \\
    \midrule
    Crop \& Flip & $\times$ & 70.18 \\
    \midrule
    Rotate \& Blur & $\times$ & 74.77 \\
    \midrule
    Crop \& Blur & $\times$ & 76.15 \\
    \midrule
    Blur & \checkmark & \textbf{82.11} \\
    \bottomrule
    \end{tabular}
    }
    \caption{
    \textbf{Ablation of complex visual perturbation considering combinations of various techniques.}  
    We observe that blurring, as a semantic-equivalent perturbation, shows optimal results. The combination of semantic-inequivalent and semantic-equivalent perturbations still results in inequivalent perturbations. Consequently, these settings yield suboptimal results.
    }
    \label{tab:supp_complex_visual_perturbation}
\end{table}

\section{More Ablation Studies}
\noindent \textbf{Combination order for visual and textual prompts.} We present the ablation study of the combination order for vision-language prompts in Table~\ref{tab:supp_perturbation_pairing}. For `Random Shuffle', we randomly shuffle the order of textual prompts to achieve a random combination of visual and textual prompts. $I_j$ exhibits a higher degree of perturbation than $I_i$ when $j > i$. Similarly, $T_j$ exhibits a higher degree of perturbation than $T_i$ when $j > i$. We observe that pairing prompts from both modalities in a similar order achieves the best detection performance. This approach generates prompt pairs with progressively increasing degrees of perturbation, effectively eliciting LVLM uncertainty.

\begin{table*}[!t]
    \centering
    \resizebox{\linewidth}{!}{
    \begin{tabular}{c|c|c}
    \toprule
    \textbf{Text Perturtbation Instruction} & \textbf{Detailed} & \textbf{Hallucination Detection Accuraccy}  \\
    \midrule
    Please rephrase it. & $\times$ & 72.48 \\
    \midrule
    Rephrase it by altering the wording, structure, or grammar, while keeping the meaning intact. & \checkmark & 75.23 \\
    \midrule
    Rewrite it using different words, sentence structure, or narrative style. Make sure the rephrased question has the same semantic meaning. & \checkmark & 77.06 \\
    \midrule
    Modify it by changing the wording, sentence flow, or grammatical structure, ensuring the meaning remains unchanged. & \checkmark & 78.90  \\
    \midrule
    Generate a \textcolor{myred}{\textit{semantic-equivalent}} variation by changing the wording, structure, grammar, or narrative. & \checkmark & 82.11 \\
    \bottomrule
    \end{tabular}
    }
    \caption{
    \textbf{Ablation study of different textual perturbation instructions.}  
    Instructions explicitly mentioning `semantic-equivalent' outperform all others, validating the efficacy of emphasizing semantic equivalence during textual perturbation. We observe that instructing LLMs with more detailed requirements yields better results. In contrast, simply asking the LLM to rephrase the original question does not produce satisfactory results. This highlights the importance of properly designed instructions.
    }
    \label{tab:supp_llm_perturbation_prompt}
\end{table*}

\begin{table}[!t]
    \centering
    \resizebox{\linewidth}{!}{
    \begin{tabular}{c|c|c}
    \toprule
    \textbf{Textual Perturbation} & \textbf{Semantic-equivalent} & \textbf{Hallucination Detection Accuracy} \\
    \midrule
    Text Shuffle & $\times$ & 71.56 \\
    \midrule
    Noise Injection & $\times$ & 73.85 \\
    \midrule
    Word Dropout & $\times$ & 74.31 \\
    \midrule
    Character Dropout & $\times$ & 72.48 \\
    \midrule
    LLM Rephrasing & \checkmark & \textbf{82.11} \\
    \bottomrule
    \end{tabular}
    }
    \caption{
    \textbf{Ablation of more textual perturbation.}
    LLM rephrasing that ensures the semantic equivalence of the original question achieves the best results. Other text perturbation techniques that do not explicitly ensure semantic equivalence yield suboptimal results.
    }
    \label{tab:supp_other_textual_perturbation}
\end{table}

\noindent \textbf{Complex visual perturbations.} We also present the ablation results of complex visual perturbations in Table~\ref{tab:supp_complex_visual_perturbation}. For `rotation and shift', we first rotate the original image by degrees in [-40, -20, 10, 20, 40], then shift the image upward by 100 pixels. For `crop and flip', we first crop the original image to ratios in [0.95, 0.9, 0.85, 0.8, 0.75], then flip the image along the horizontal axis. For `rotate and blur', we first apply rotation by degrees in [-40, -20, 10, 20, 40], then blur the image with a radius of 1. For `crop and blur', we first crop the original image to ratios in [0.95, 0.9, 0.85, 0.8, 0.75], then blur the image with a radius of 1. We observe that blurring still achieves optimal results compared to those complex combined augmentations. Combining both semantic-equivalent and -inequivalent perturbations can disrupt the semantic-equivalent property of the visual prompt, yielding suboptimal results.

\noindent \textbf{Various textual perturbation instructions.} We report the ablation study of LLM instructions for textual perturbation in Table~\ref{tab:supp_llm_perturbation_prompt}. We observe that detailed prompts explicitly mentioning `semantic-equivalent' yield the best results. In contrast, prompts with simple instructions do not produce satisfactory results. Moreover, using detailed instructions with specific requirements is beneficial for achieving higher detection accuracy.

\noindent \textbf{Different textual perturbations.} We also present the ablation study of various textual prompt perturbations in Table~\ref{tab:supp_other_textual_perturbation}. For `text shuffle', we randomly shuffle the original question at the word level. For `noise injection', we randomly inject nonsensical words and characters into the original questions. For `word dropout', we drop words from the original text with a ratio of 0.2. For `character dropout', we randomly remove characters with a dropout rate of 0.1. LLM rephrasing surpasses all other semantic-inequivalent perturbations. Other basic approaches that do not consider semantic preservation yield suboptimal results.

\begin{table}[!t]
    \centering
    \resizebox{\linewidth}{!}{
    \begin{tabular}{c|c|c|c}
    \toprule
    \textbf{LVLM Group} & \textbf{$\#$Param.} & \textbf{Source} & \textbf{Hugging Face Version} \\
    \midrule
    Qwen2VL & 72B & Alibaba & Qwen/Qwen2-VL-72B-Instruct \\
    \midrule
    Qwen2VL & 7B & Alibaba & Qwen/Qwen2-VL-7B-Instruct \\
    \midrule
    Qwen2VL & 2B & Alibaba & Qwen/Qwen2-VL-2B-Instruct \\
    \midrule
    InternVL2 & 26B & OpenGVLab & OpenGVLab/InternVL2-26B \\
    \midrule
    InternVL2 & 8B & OpenGVLab & OpenGVLab/InternVL2-8B \\
    \midrule
    InternVL2 & 1B & OpenGVLab & OpenGVLab/InternVL2-1B \\
    \midrule
    LLaVANeXT & 13B & UW–Madison & llava-hf/llava-v1.6-vicuna-13b-hf \\
    \midrule
    LLaVANeXT & 7B & UW–Madison & llava-hf/llava-v1.6-mistral-7b-hf \\
    \midrule
    LLaVA1.5 & 13B & UW–Madison & llava-hf/llava-1.5-13b-hf \\
    \midrule
    LLaVA1.5 & 7B & UW–Madison & llava-hf/llava-1.5-7b-hf \\
    \bottomrule
    \end{tabular}
    }
    \caption{
    \textbf{Detailed information about adopted LVLMs.}  
    We utilize 10 different LVLMs for hallucination detection experiments, with model capacities ranging from 1B (InternVL2-1B) to 72B (Qwen2VL-72B). The LVLMs come from diverse development backgrounds, including university research and industry collaborations. `$\#$Param.' refers to the total number of parameters in each LVLM.
    }
    \label{tab:supp_LVLM}
\end{table}

\section{More Qualitative Results}

\begin{figure*}
    \centering
    \includegraphics[width=0.9\linewidth]{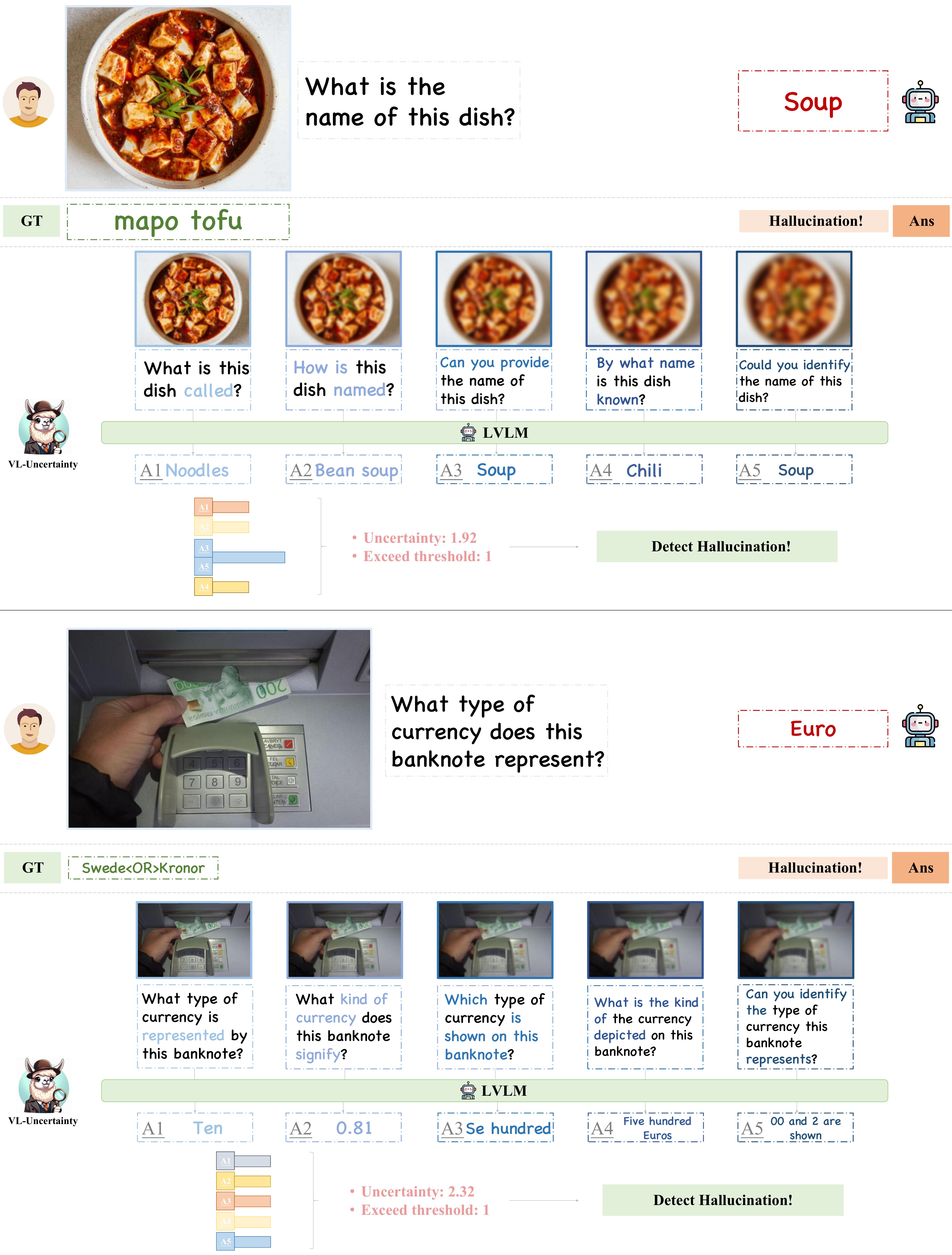}
    \caption{
    Successful hallucination detection cases from VL-Uncertainty on MMVet.
    }
    \label{fig:supp_vis_MMVet}
\end{figure*}

\begin{figure*}
    \centering
    \includegraphics[width=\linewidth]{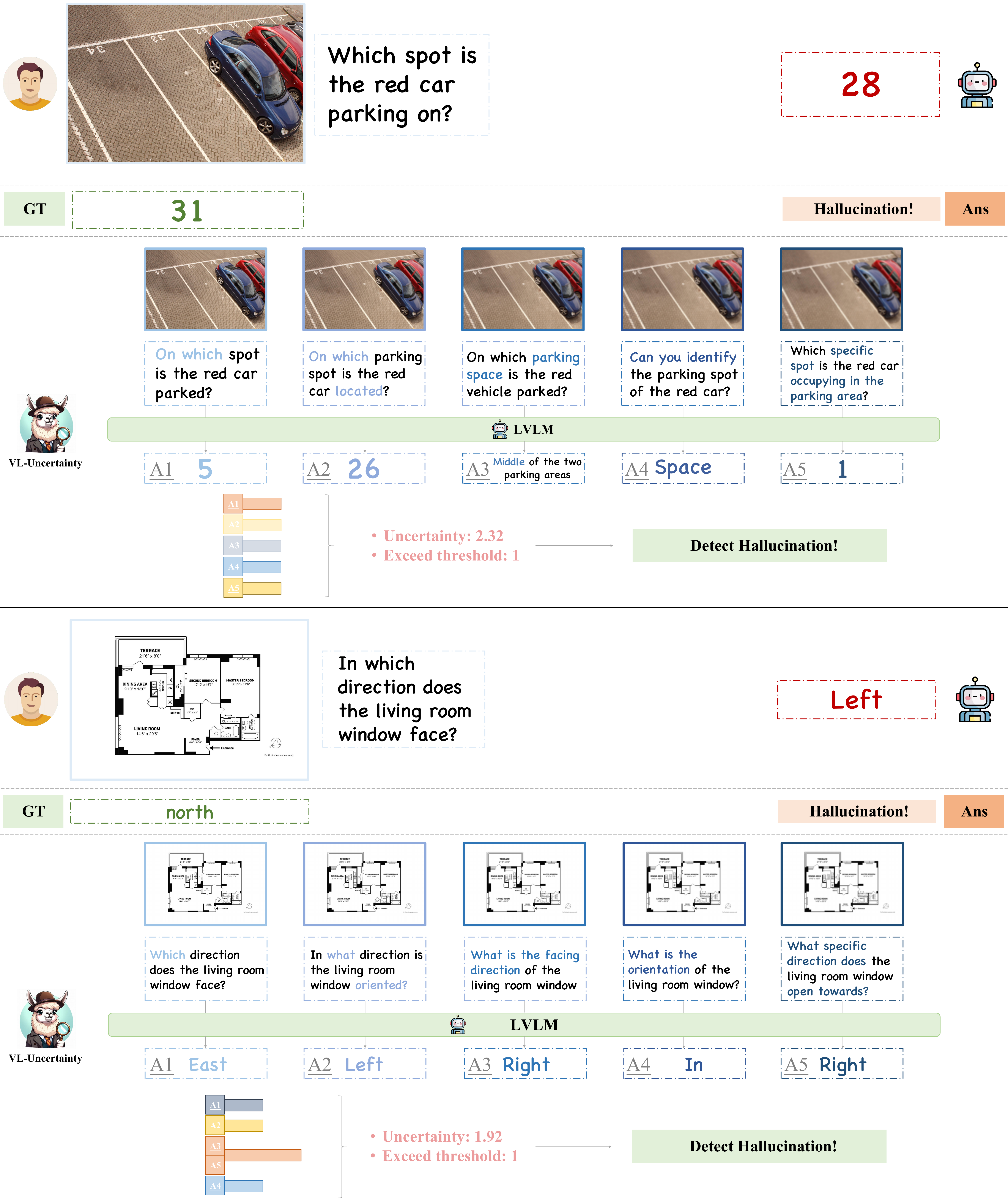}
    \caption{
    Successful hallucination detection cases from VL-Uncertainty on MMVet.
    }
    \label{fig:supp_vis_MMVet2}
\end{figure*}

\begin{figure*}
    \centering
    \includegraphics[width=\linewidth]{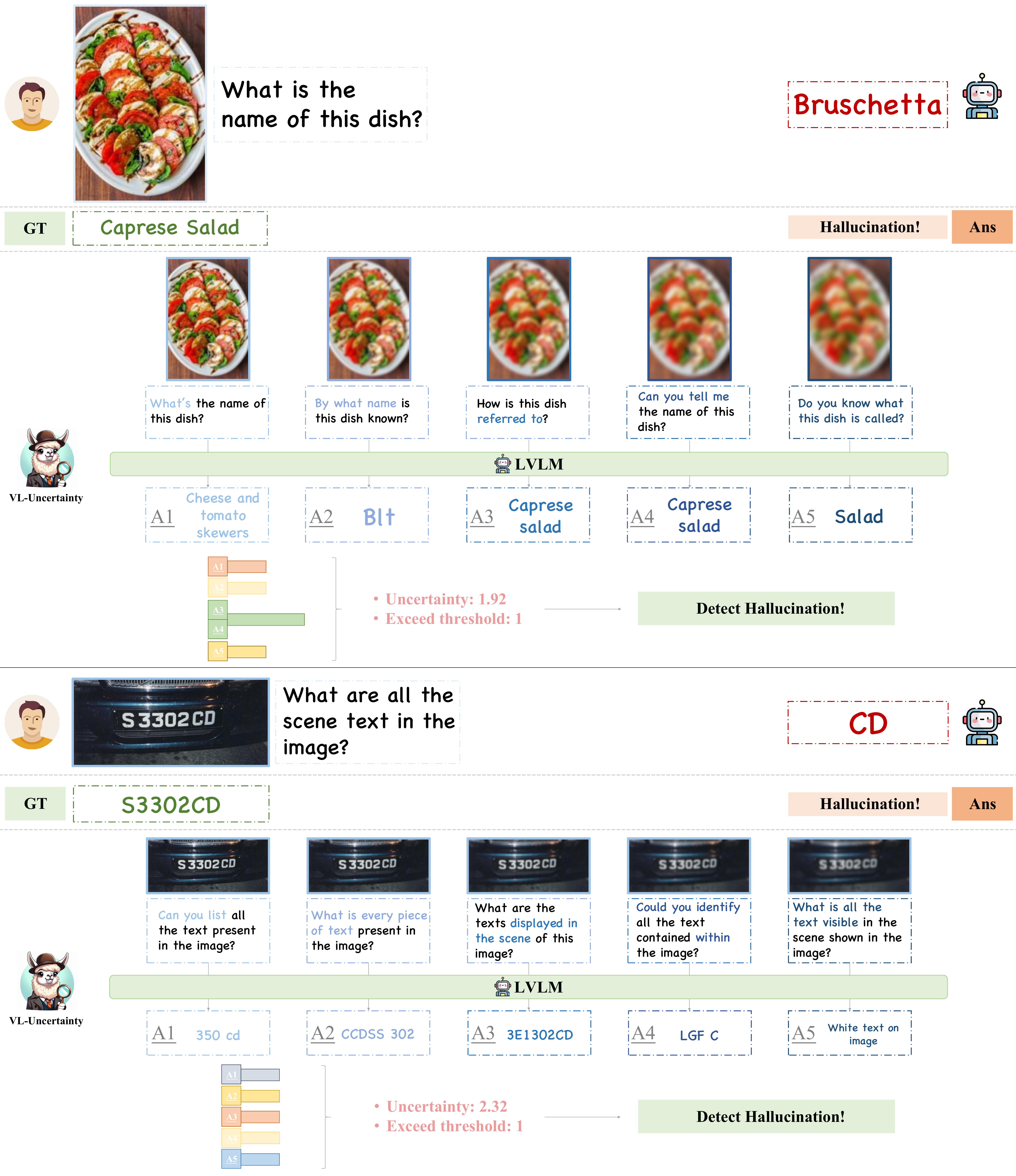}
    \caption{
    Successful hallucination detection cases from VL-Uncertainty on MMVet.
    }
    \label{fig:supp_vis_MMVet3}
\end{figure*}

\noindent \textbf{On MMVet.} We present case studies of accurate VL-Uncertainty hallucination detection on MMVet (free-form benchmark) in Fig.~\ref{fig:supp_vis_MMVet}, Fig.~\ref{fig:supp_vis_MMVet2}, and Fig.~\ref{fig:supp_vis_MMVet3}.  

\begin{figure*}
    \centering
    \includegraphics[width=\linewidth]{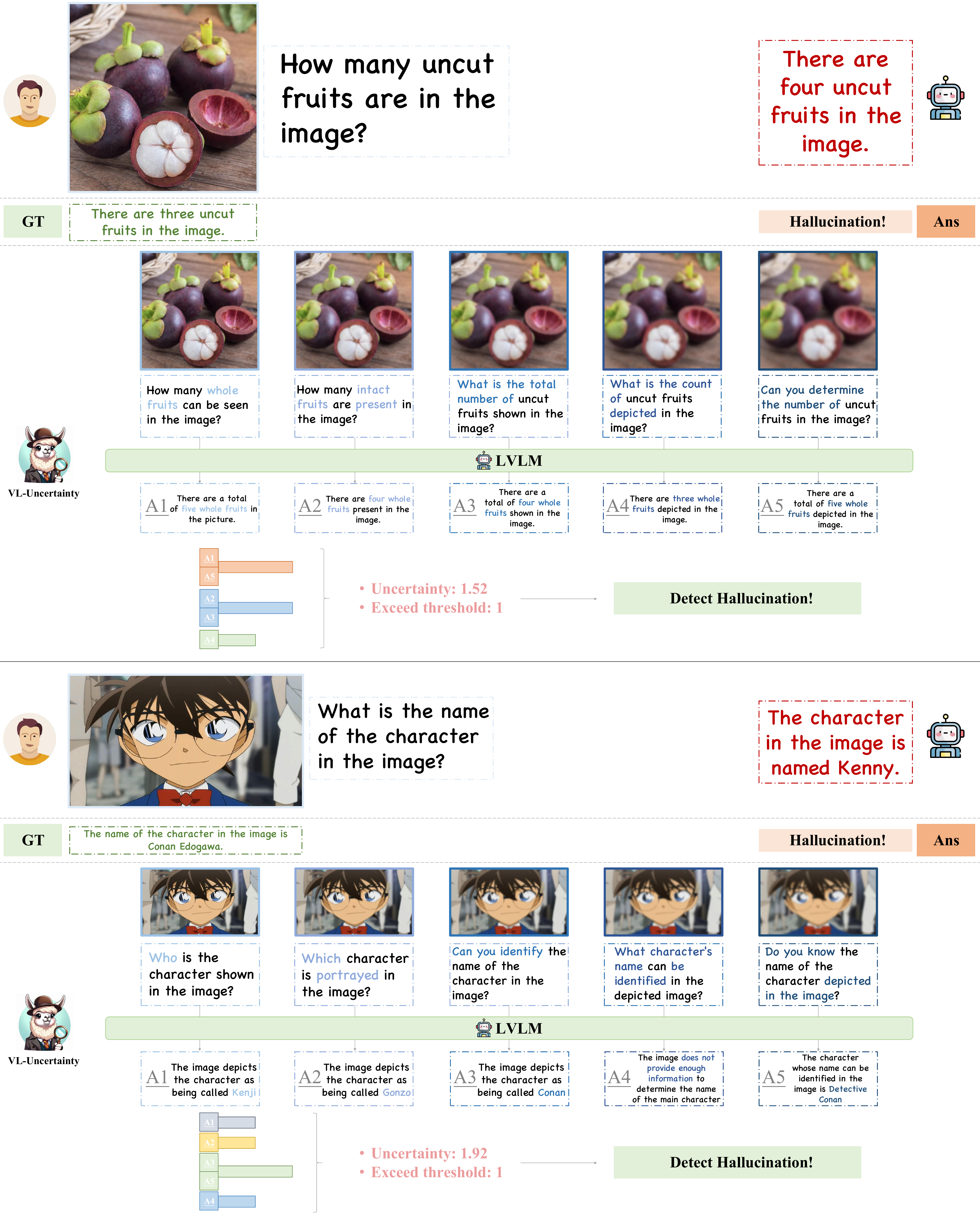}
    \caption{
    Successful hallucination detection cases from VL-Uncertainty on LLaVABench.
    }
    \label{fig:supp_vis_LLaVABench}
\end{figure*}

\begin{figure*}
    \centering
    \includegraphics[width=0.9\linewidth]{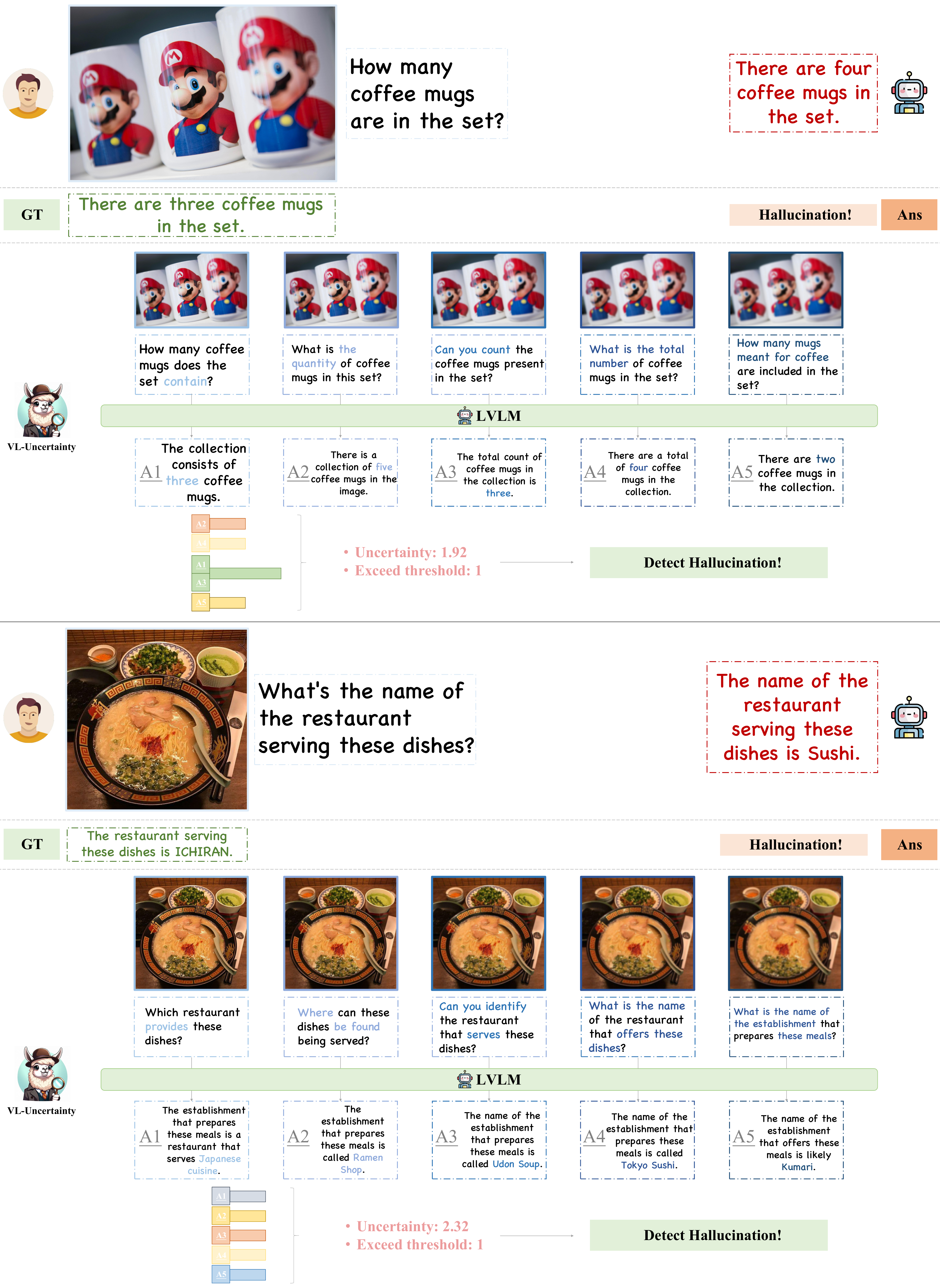}
    \caption{
    Successful hallucination detection cases from VL-Uncertainty on LLaVABench.
    }
    \label{fig:supp_vis_LLaVABench2}
\end{figure*}

\begin{figure*}
    \centering
    \includegraphics[width=0.9\linewidth]{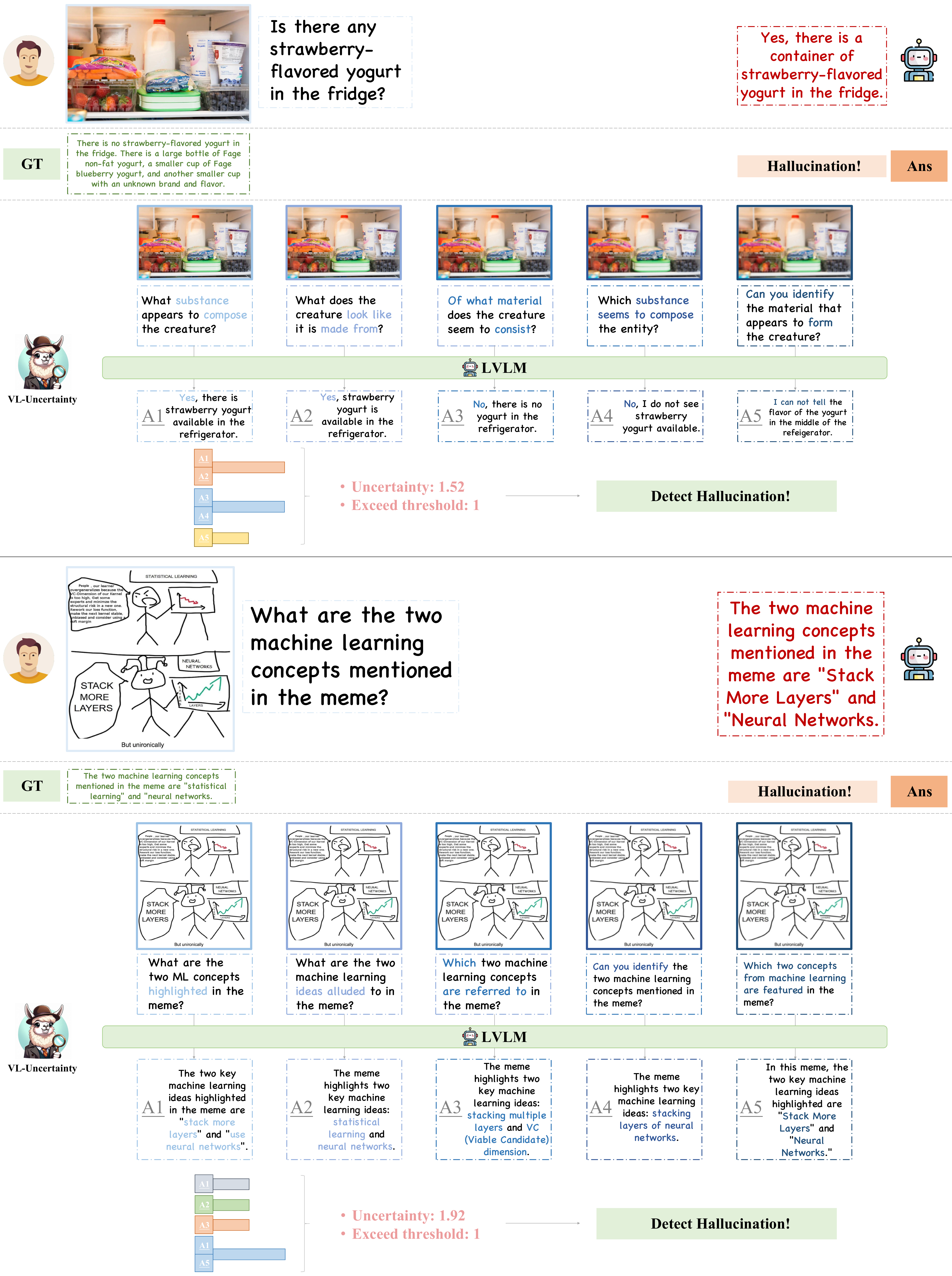}
    \caption{
    Successful hallucination detection cases from VL-Uncertainty on LLaVABench.
    }
    \label{fig:supp_vis_LLaVABench3}
\end{figure*}

\noindent \textbf{On LLaVABench.} We also show successful detection cases from LLaVABench with VL-Uncertainty in Fig.~\ref{fig:supp_vis_LLaVABench}, Fig.~\ref{fig:supp_vis_LLaVABench2}, and Fig.~\ref{fig:supp_vis_LLaVABench3}.  

\begin{figure*}
    \centering
    \includegraphics[width=\linewidth]{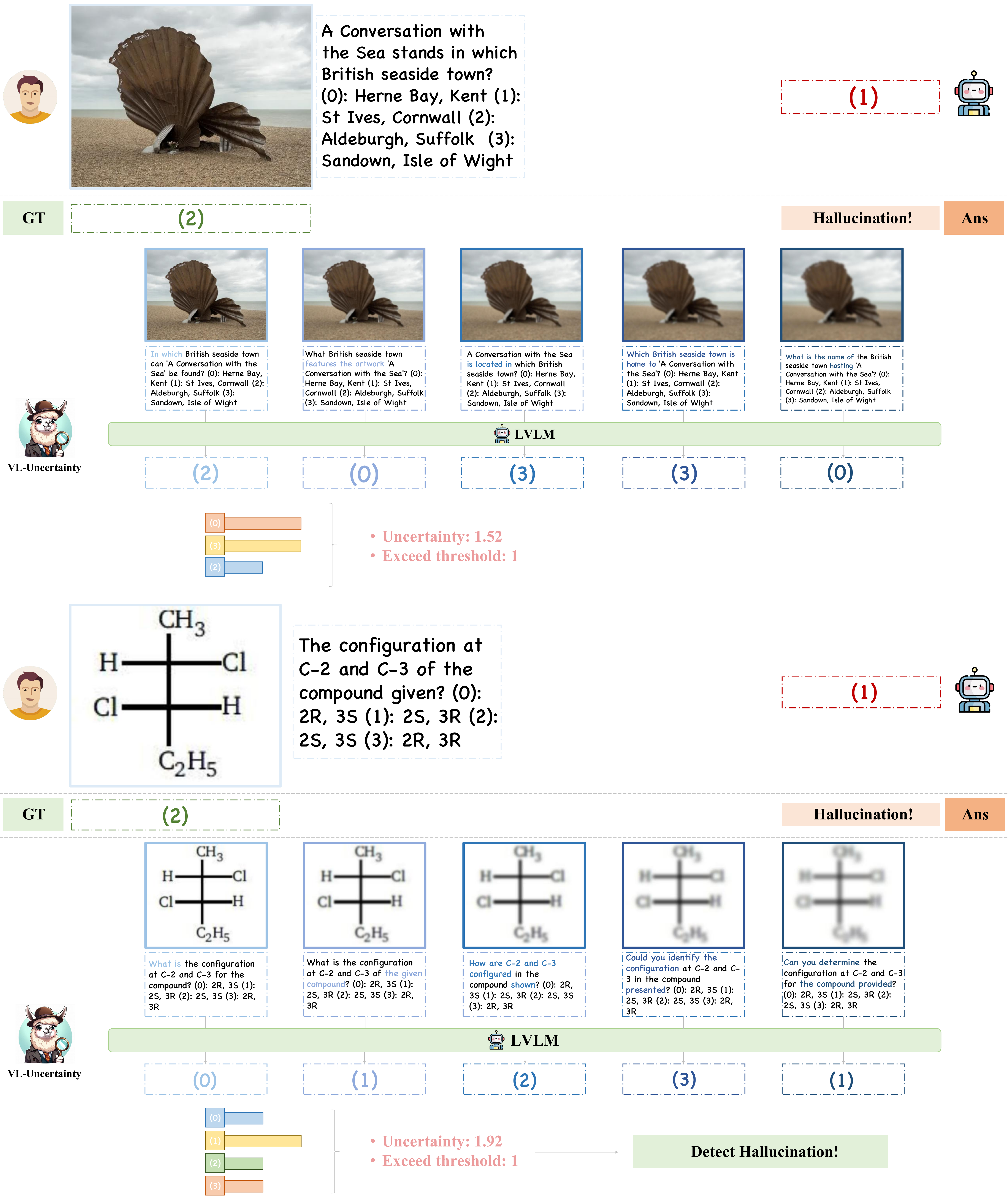}
    \caption{
    Successful hallucination detection cases from VL-Uncertainty on MMMU.
    }
    \label{fig:supp_vis_MMMU}
\end{figure*}

\begin{figure*}
    \centering
    \includegraphics[width=\linewidth]{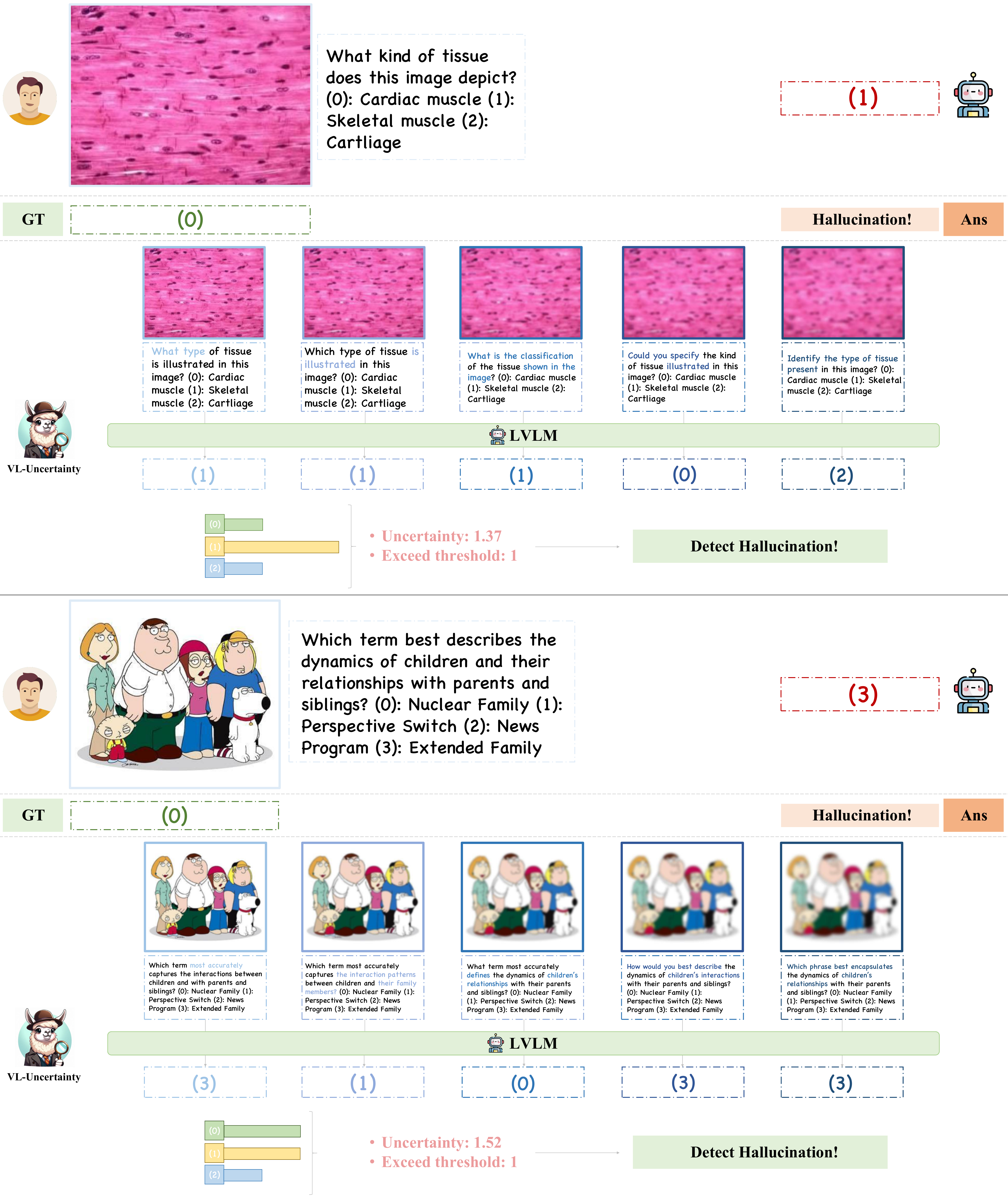}
    \caption{
    Successful hallucination detection cases from VL-Uncertainty on MMMU.
    }
    \label{fig:supp_vis_MMMU2}
\end{figure*}

\begin{figure*}
    \centering
    \includegraphics[width=\linewidth]{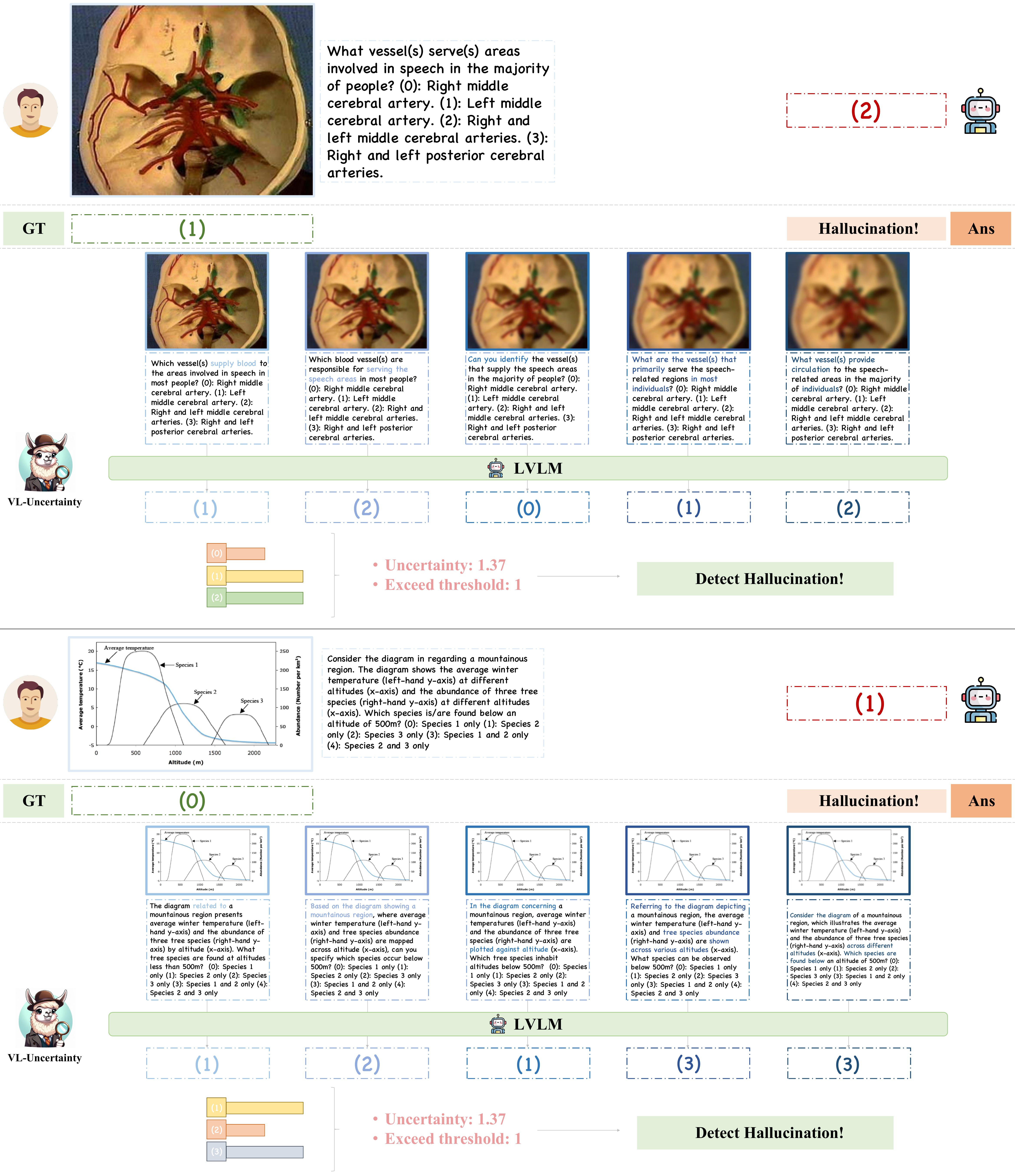}
    \caption{
    Successful hallucination detection cases from VL-Uncertainty on MMMU.
    }
    \label{fig:supp_vis_MMMU3}
\end{figure*}

\noindent \textbf{On MMMU.} We also provide case studies of accurate hallucination detection from VL-Uncertainty on MMMU (multi-choice benchmark) in Fig.~\ref{fig:supp_vis_MMMU}, Fig.~\ref{fig:supp_vis_MMMU2}, and Fig.~\ref{fig:supp_vis_MMMU3}.

\begin{figure*}
    \centering
    \includegraphics[width=\linewidth]{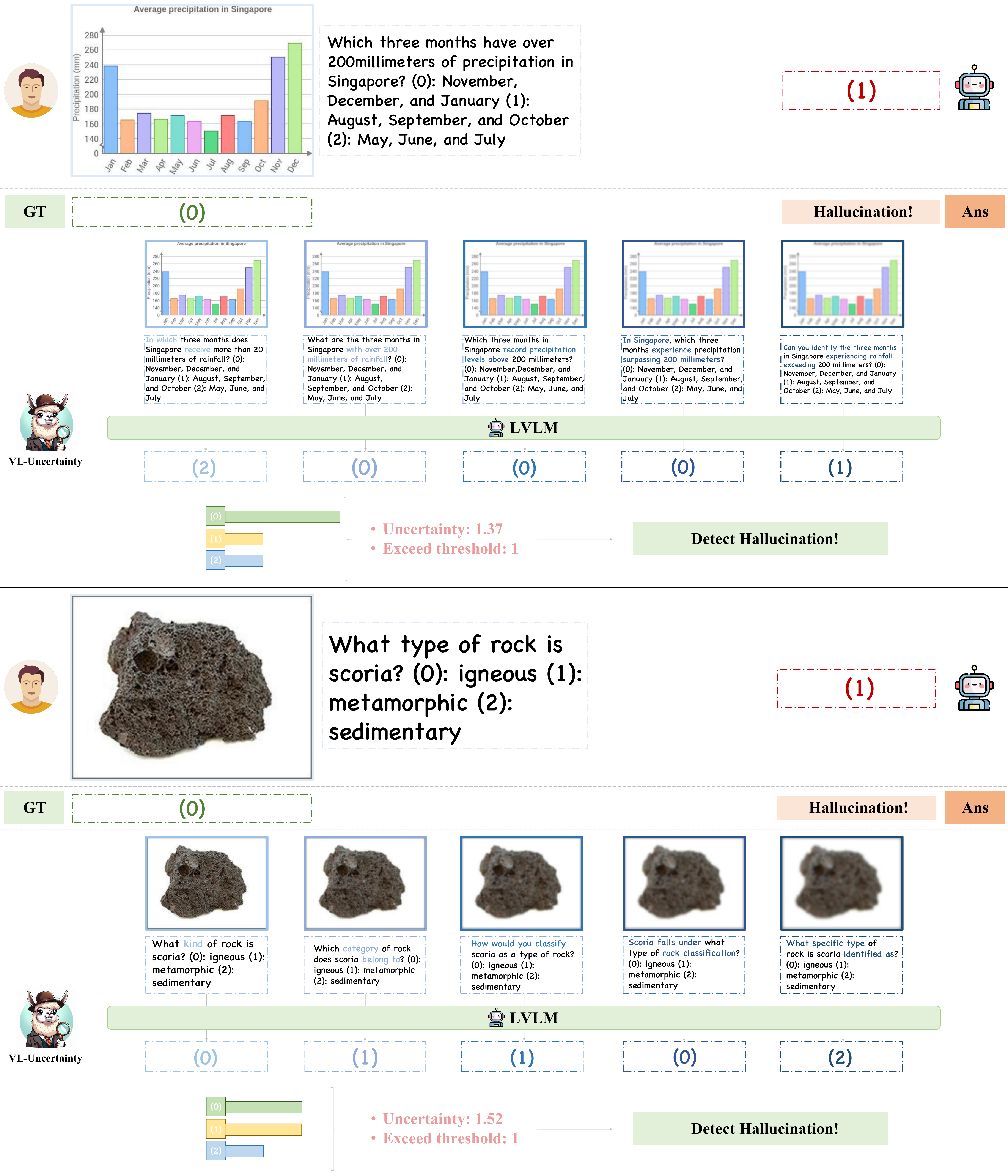}
    \caption{
    Successful hallucination detection cases from VL-Uncertainty on ScienceQA.
    }
    \label{fig:supp_vis_ScienceQA}
\end{figure*}

\begin{figure*}
    \centering
    \includegraphics[width=\linewidth]{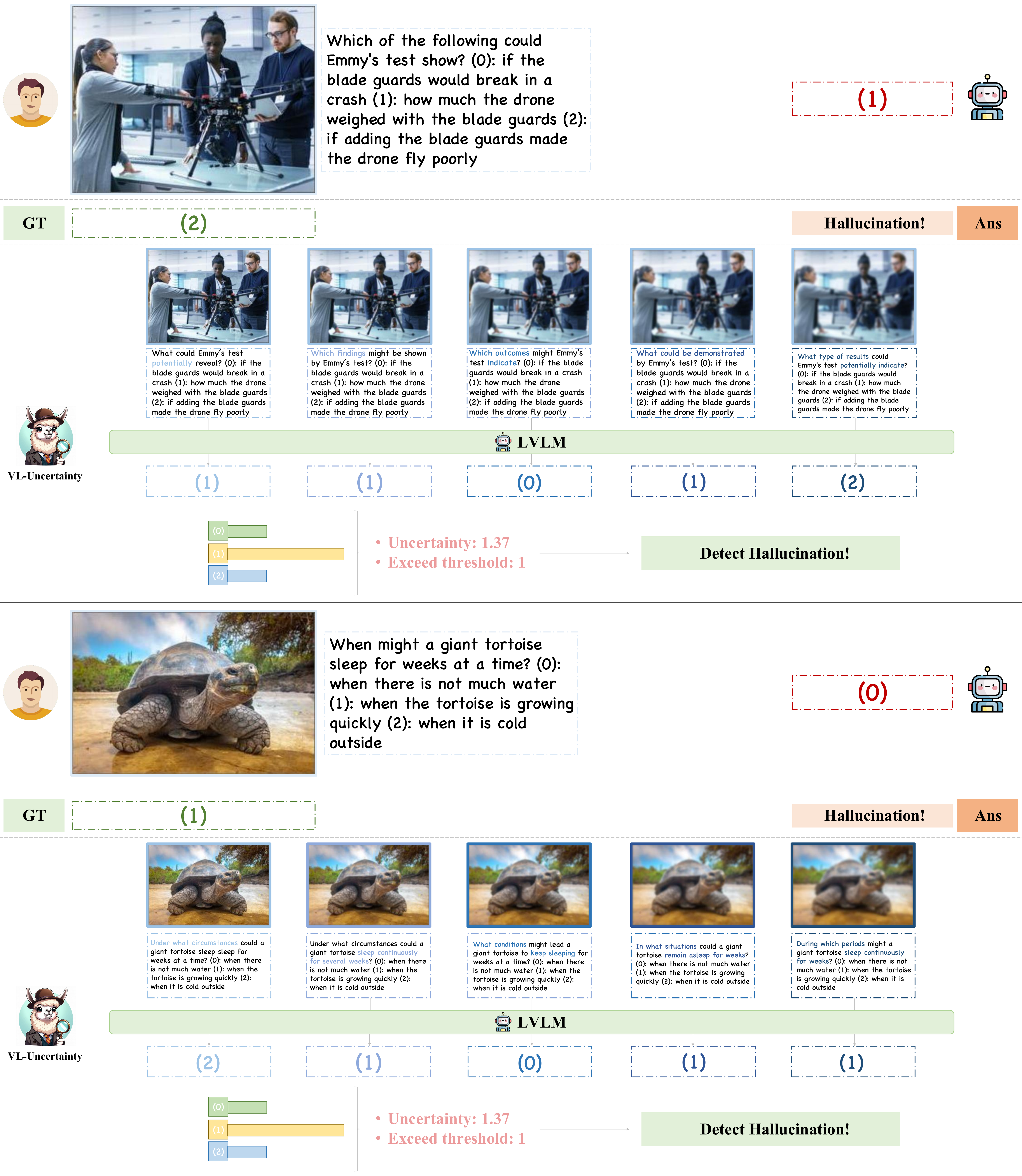}
    \caption{
    Successful hallucination detection cases from VL-Uncertainty on ScienceQA.
    }
    \label{fig:supp_vis_ScienceQA2}
\end{figure*}

\begin{figure*}
    \centering
    \includegraphics[width=\linewidth]{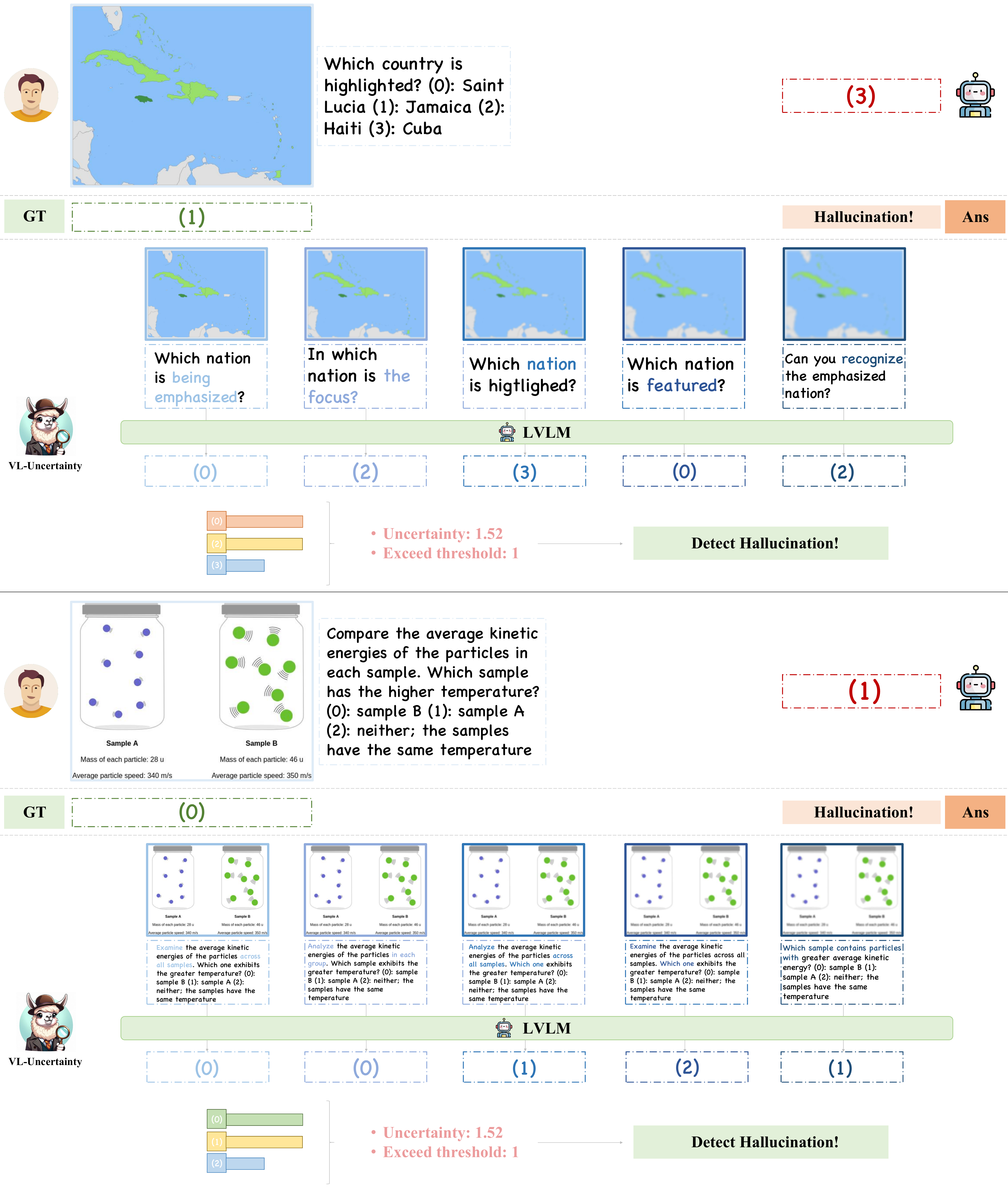}
    \caption{
    Successful hallucination detection cases from VL-Uncertainty on ScienceQA.
    }
    \label{fig:supp_vis_ScienceQA3}
\end{figure*}

\noindent \textbf{On ScienceQA.} We observe VL-Uncertainty can also achieve accurate hallucination detection on complex multi-choice questions (see Fig.~\ref{fig:supp_vis_ScienceQA}, Fig.~\ref{fig:supp_vis_ScienceQA2}, and  Fig.~\ref{fig:supp_vis_ScienceQA3}).

\begin{figure*}
    \centering
    \includegraphics[width=\linewidth]{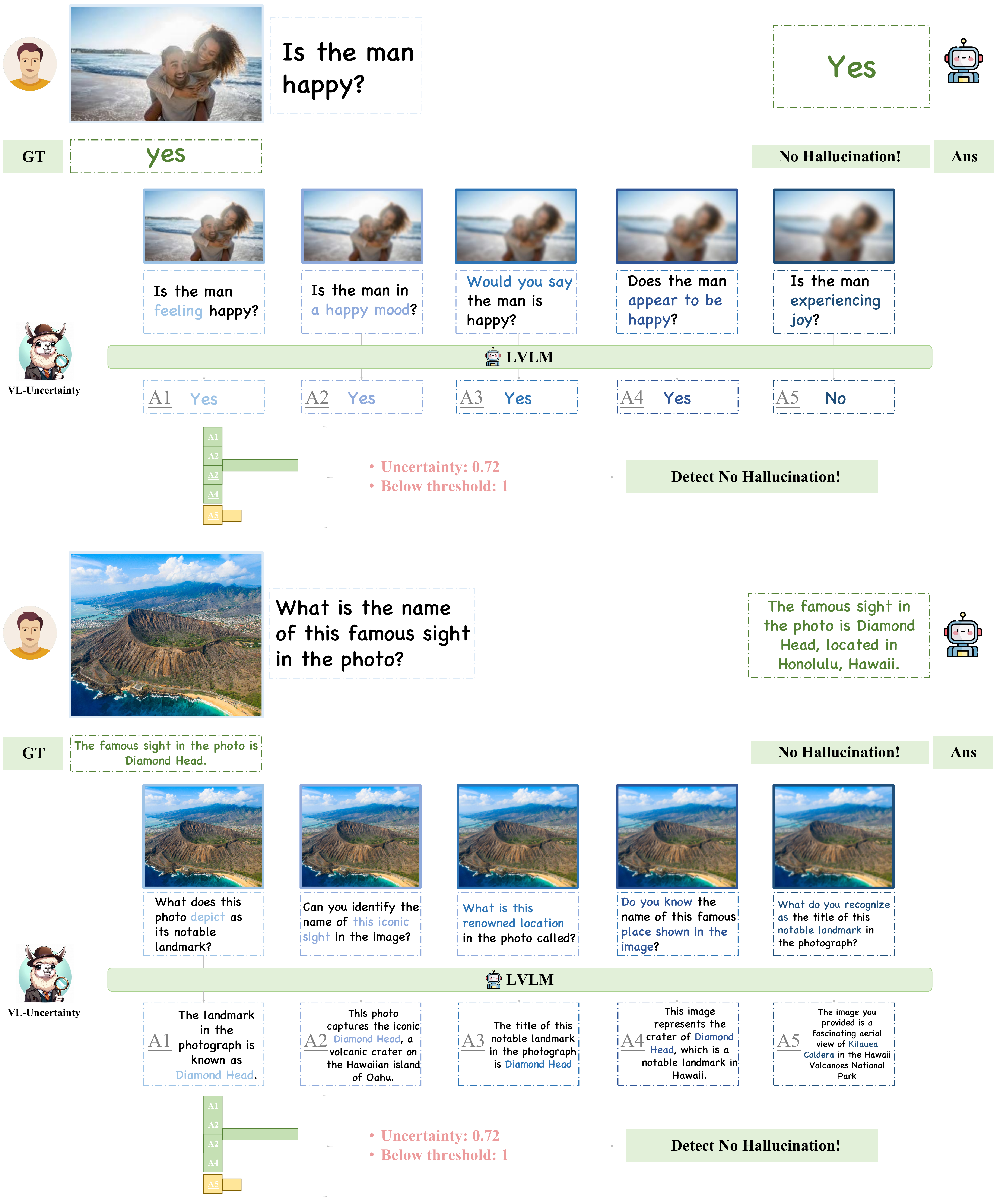}
    \caption{
    VL-Uncertainty can also accurately detect correct answers as non-hallucinatory. These are two free-form cases from MMVet and LLaVABench.
    }
    \label{fig:supp_vis_MMVet_LLaVABench_correct}
\end{figure*}

\begin{figure*}
    \centering
    \includegraphics[width=\linewidth]{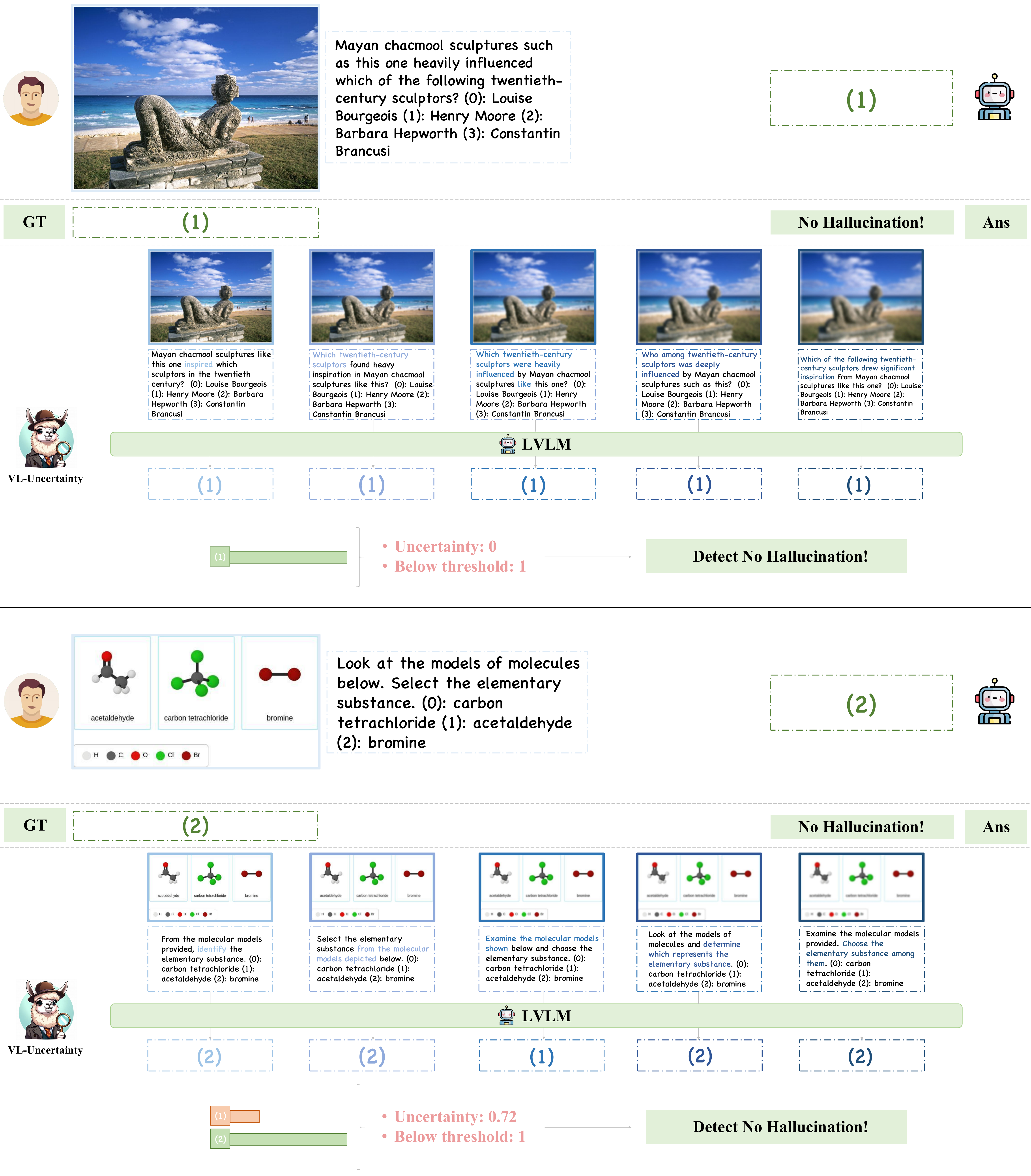}
    \caption{
    VL-Uncertainty can also accurately detect correct answers as non-hallucinatory. These are two multi-choice cases from MMMU and ScienceQA.
    }
    \label{fig:supp_vis_MMMU_ScienceQA_correct}
\end{figure*}

\noindent \textbf{Non-hallucinatory answer detection.} VL-Uncertainty is also capable of accurately identifying correct answers as non-hallucinatory. We showcase successful cases from free-form benchmarks (see Fig.~\ref{fig:supp_vis_MMVet_LLaVABench_correct}) and multi-choice benchmarks (see Fig.~\ref{fig:supp_vis_MMMU_ScienceQA_correct}).

\begin{table*}[!t]
    \centering
    \resizebox{\textwidth}{!}{
    \begin{tabular}{c|c|c|c|c|c|c|c|c|c|c}
    \toprule
    \textbf{LVLM Group} & \textbf{Do Sample} & \textbf{Temperature} & \textbf{Repetition Penalty} & \textbf{Top K} & \textbf{Top P} & \textbf{Max New Tokens} & \textbf{Answer Prefix} & \textbf{Flash Attention} & \textbf{Data Type} & \textbf{Visual Preprocess}  \\
    \midrule
    Qwen2VL & \checkmark & 0.1 & 1.05 & 50 & 0.95 & 32 &  - & \checkmark & torch.bfloat16 & qwen util \\
    \midrule
    InternVL2 & \checkmark & 0.1 & - & - & - & 32 & - & \checkmark & torch.bfloat16 & resize \\
    \midrule
    LLaVA1.5 & \checkmark & 0.1 & - & - & - & 32 & ASSISTANT: & \checkmark & torch.bfloat16 & - \\
    \midrule
    LLaVANeXT & \checkmark & 0.1 & - & - & - & 32 & [/INST] / ASSISTANT: & \checkmark & torch.bfloat16 & - \\
    \bottomrule
    \end{tabular}
    }
    \caption{
    \textbf{Detailed hyper-parameter settings adopted for LVLMs.}  
    All LVLMs utilize flash attention to optimize time and memory efficiency. Sampling is also enabled for all LVLMs to facilitate the uncertainty estimation process.
    }
    \label{tab:supp_LVLM_hp}
\end{table*}

\begin{table*}[!t]
    \centering
    \resizebox{\textwidth}{!}{
    \begin{tabular}{c|c|c|c|c|c|c|c}
    \toprule
    \textbf{Benchmark} & \textbf{Year} & \textbf{Conference} & \textbf{Source} & \textbf{Hugging Face Version} & \textbf{Format} & \textbf{Size Before Filtering} & \textbf{Size After Filtering} \\
    \midrule
    MMMU & 2024 & CVPR (Oral) & OSU & MMMU/MMMU & Multi-choice & 900 & 825 \\
    \midrule
    MM-Vet & 2023 & ICML & NUS & whyu/mm-vet & Free-form & 218 & 218 \\
    \midrule
    LLaVABench & 2023 & NeurIPS (Oral) & UW–Madison & lmms-lab/llava-bench-in-the-wild & Free-form & 60 & 60 \\
    \midrule
    ScienceQA & 2022 & NeurIPS & UCLA & derek-thomas/ScienceQA & Multi-choice & 4241 & 2017 \\
    \bottomrule
    \end{tabular}
    }
    \caption{
    \textbf{Detailed information about adopted benchmarks.}  
    We utilize four different benchmarks for hallucination detection experiments. Both free-form benchmarks (MM-Vet and LLaVABench) and multiple-choice benchmarks (MMMU and ScienceQA) are included. All these benchmarks provide a multi-modal context, \emph{e.g.}, both visual and textual prompts are supplied.
    }
    \label{tab:supp_benchmark}
\end{table*}

\begin{table*}[!t]
    \centering
    \resizebox{\linewidth}{!}{
    \begin{tabular}{c|c|c|c}
    \toprule
    \textbf{Benchmark} & \textbf{Format} & \textbf{Added Prompt} & \textbf{Choice Marker} \\
    \midrule
    ScienceQA & Multi-Choice & This is a single choice question, answer only with choice number in \{choice\_numbers\}. & 0,1,... \\
    \midrule
    MMMU & Multi-Choice & This is a single choice question, answer only with choice number in \{choice\_numbers\}. & 0,1,... \\
    \midrule
    MM-Vet & Free-Form & NOTE: Provide only the final answer. Do not provide unrelated details. & - \\ 
    \midrule
    LLaVABench & Free-Form & - & - \\
    \bottomrule
    \end{tabular}
    }
    \caption{
    \textbf{Implementation details for utilized benchmarks.}  
    Both ScienceQA and MMMU contain multiple-choice questions, where the context and choice lists are provided, and the ground truth answer is represented by a choice number. In contrast, MM-Vet and LLaVABench contain free-form questions without choice lists, and the ground truth answers can vary in length and format. For the multiple-choice benchmarks, we convert all choice markers to 0, 1, ..., and prompt the LVLM to respond with a single number only. For MM-Vet, we prompt the LVLM to provide only the final answer and ignore unrelated details, as the ground truth answers for MM-Vet are relatively concise.
    }
    \label{tab:supp_benchmark_detail}
\end{table*}

\begin{table}[!t]
    \centering
    \resizebox{\linewidth}{!}{
    \begin{tabular}{c|c|c|c|c}
    \toprule
    \textbf{LLM Group} & \textbf{$\#$Param.} & \textbf{Source} & \textbf{Hugging Face Version} & \textbf{Multi-GPU Support} \\
    \midrule
    Qwen2.5 & 0.5B & Alibaba & Qwen/Qwen2.5-0.5B-Instruct & \checkmark \\
    \midrule
    Qwen2.5 & 1.5B & Alibaba & Qwen/Qwen2.5-1.5B-Instruct & \checkmark \\
    \midrule
    Qwen2.5 & 3B & Alibaba & Qwen/Qwen2.5-3B-Instruct & \checkmark \\
    \midrule
    Qwen2.5 & 7B & Alibaba & Qwen/Qwen2.5-7B-Instruct & \checkmark \\
    \bottomrule
    \end{tabular}
    }
    \caption{
    \textbf{Adopted LLM details.} We adopt Qwen2.5 as the LLM for all our experiments. The LLM is utilized for: (1) textual semantic-equivalent perturbation: The LLM is prompted to rephrase the original questions to various extents. (2) semantic clustering: The LLM checks whether Answer A and Answer B entail each other. It is prompted to check entailment bidirectionally, \emph{e.g.}, `If A entails B' and `If B entails A'. Entailment is identified only when bidirectional entailment is satisfied. (3) free-form question evaluation: The LLM verifies whether the LVLM's answers to free-form questions are correct, as rule-based metrics are unsuitable for free-form scenarios.
    }
    \label{tab:supp_llm}
\end{table}

\begin{table}[!t]
    \centering
    \resizebox{\linewidth}{!}{
    \begin{tabular}{c|c|c|c|c}
    \toprule
    \textbf{Do Sample} & \textbf{Temperature} & \textbf{Repetition Penalty} & \textbf{Top P} & \textbf{Max New Tokens} \\
    \midrule
    \checkmark & 0.1 & 1.05 & 0.8 & 256 \\
    \bottomrule
    \end{tabular}
    }
    \caption{
    \textbf{Hyper-parameters for LLM.}
    We adopt the same settings for all Qwen2.5 LLMs (0.5B, 1.5B, 3B, 7B). A low temperature (0.1) is used during LVLM answer checking and answer entailment checking. The temperature is increased during textual perturbation to facilitate rephrasing the original questions to varying degrees.
    }
    \label{tab:supp_llm_hp}
\end{table}

\begin{figure*}
    \centering
    \includegraphics[width=\linewidth]{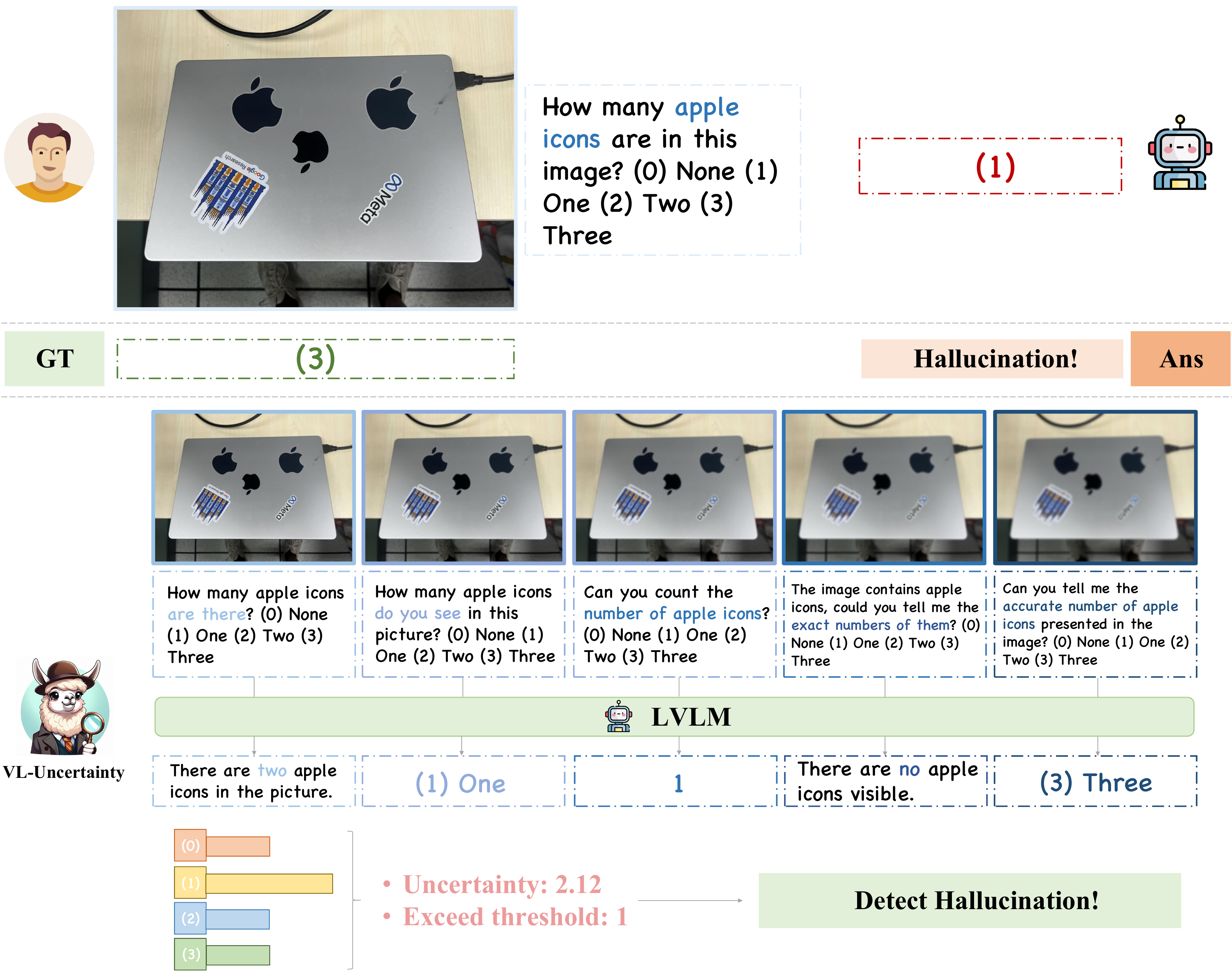}
    \caption{
    \textbf{Physical-world cases for multi-choice questions.}  
    VL-Uncertainty successfully detects hallucinations through semantic-equivalent perturbation and refined uncertainty estimation. This demonstrates the potential of VL-Uncertainty in physical-world applications. The picture is taken with iPhone13 in office environment, and the question is manually crafted.
    }
    \label{fig:supp_real_world_multi_choice}
\end{figure*}

\begin{figure*}
    \centering
    \includegraphics[width=\linewidth]{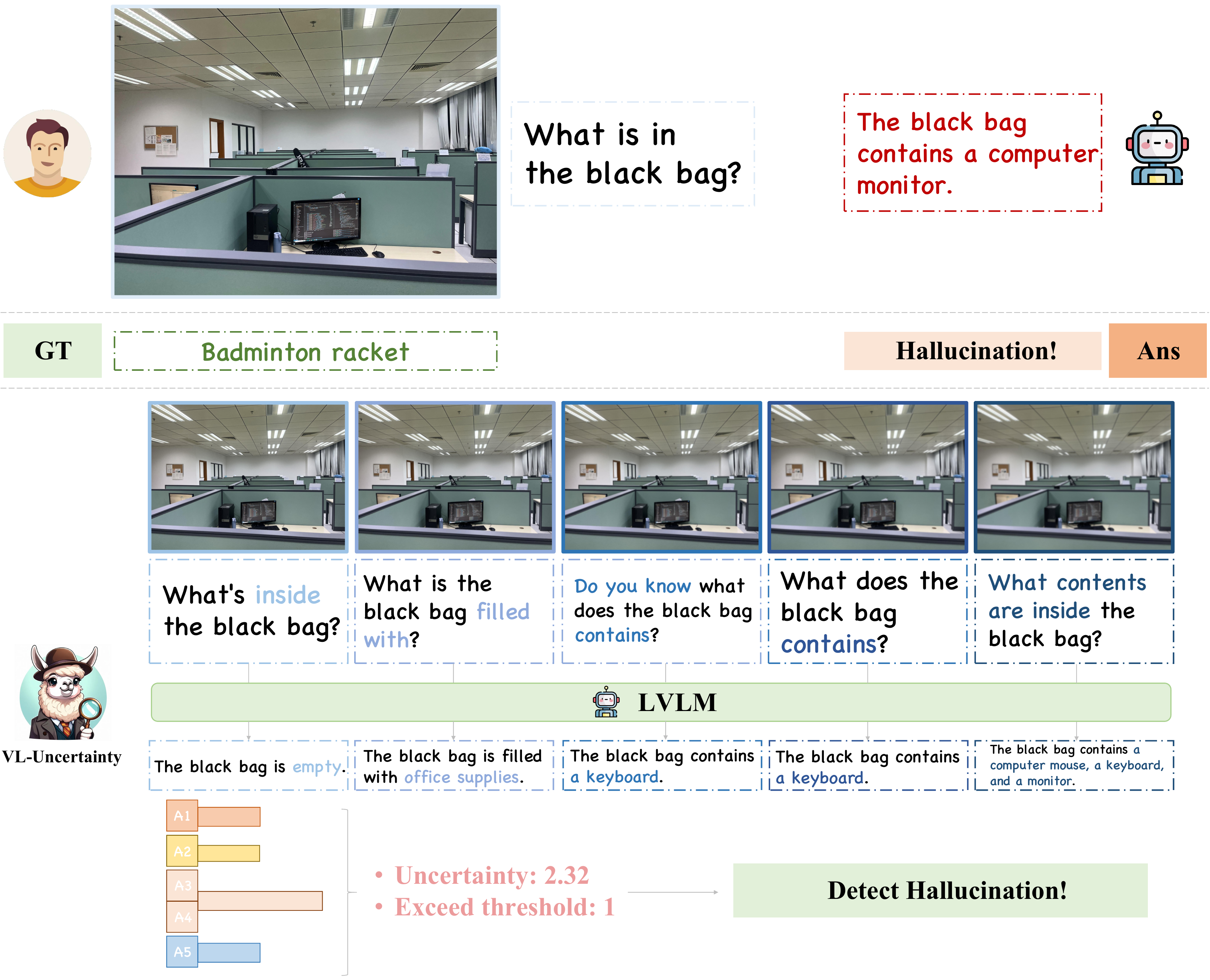}
    \caption{
    \textbf{Physical-world cases for free-form questions.}  
    For free-form questions, VL-Uncertainty also achieves accurate hallucination detection results. This validates the robustness of VL-Uncertainty across different question formats and domains. Picture is captured using iPhone13 in office environment, and the question is manually designed.
    }
    \label{fig:supp_real_world_free_form}
\end{figure*}

\noindent \textbf{Physical world cases.} We present physical-world hallucination detection scenarios in Fig.~\ref{fig:supp_real_world_free_form} (free-form questions) and Fig.~\ref{fig:supp_real_world_multi_choice} (multi-choice questions). The successful detection of hallucinations by VL-Uncertainty demonstrates its potential in related physical-world applications. Visual prompts are taken with iPhone13 in office environment and textual prompts are manually designed.

\section{More Implementation Details}

\begin{table}[!t]
    \centering
    \resizebox{\linewidth}{!}{
    \begin{tabular}{c|c|c|c}
    \toprule
    \textbf{Hugging Face Version} & \textbf{Multi-GPU} & \textbf{Total GPU Memory} & \textbf{Time Per Sample} \\
    \midrule
    OpenGVLab/InternVL2-1B & \checkmark & 4G  & 2.97s \\
    \midrule
    OpenGVLab/InternVL2-8B & \checkmark & 24G & 3.27s \\
    \midrule
    OpenGVLab/InternVL2-26B & \checkmark & 58G & 4.16s \\
    \midrule
    llava-hf/llava-1.5-7b-hf & $\times$ & 21G & 2.08s \\
    \midrule
    llava-hf/llava-1.5-13b-hf & $\times$ & 32G & 2.97s \\
    \midrule
    llava-hf/llava-v1.6-mistral-7b-hf & $\times$ & 22G & 2.97s \\
    \midrule
    llava-hf/llava-v1.6-vicuna-13b-hf & $\times$ & 35G & 4.75s \\
    \midrule
    Qwen/Qwen2-VL-2B-Instruct & \checkmark & 12G & 2.37s \\
    \midrule
    Qwen/Qwen2-VL-7B-Instruct & \checkmark & 24G & 2.78s \\
    \midrule
    Qwen/Qwen2-VL-72B-Instruct & \checkmark & 145G & 3.28s \\
    \bottomrule
    \end{tabular}
    }
    \caption{
    \textbf{GPU memory and average hallucination detection time per sample for all LVLMs.}
    Hallucination detection for most models can be realized using normal GPUs, without need for A100/H100. For average hallucination detection time per sample, the cost generally falls below 5 seconds, introducing minimal overhead. Considering the safety risks that our hallucination detection can identify and flag, the time cost is relatively efficient. All analyses are run with \textit{2 H100 (80G) GPUs}.
    }
    \label{tab:supp_cost}
    \vspace{-.2in}
\end{table}

\noindent \textbf{Utilized LVLM Details.} We present a detailed introduction to the 10 LVLMs adopted in our experiments in Table~\ref{tab:supp_LVLM}. We utilize LVLMs from four distinct model clusters, \emph{e.g.}, LLaVA1.5, LLaVANeXT, InternVL2, and Qwen2VL. The development backgrounds span academia and industry. The specific Hugging Face versions are provided to facilitate reproducibility of the results. Detailed hyper-parameters for all LVLMs are also reported (see Table~\ref{tab:supp_LVLM_hp}). For LVLMs within the same group, the adopted hyper-parameter settings remain consistent. Flash attention is enabled for all LVLMs to enhance efficiency. For Qwen2VL, additional hyper-parameters, \emph{e.g.}, `Repetition Penalty', `top-k', and `top-p', are configured to ensure diversity in responses. Otherwise, the responses remain identical.

\noindent \textbf{Utilized Benchmark details.} We also introduce four multi-modal benchmarks adopted in our study (see Table~\ref{tab:supp_benchmark}). These benchmarks are all recently released and designed to evaluate the comprehensive capabilities of LVLMs. The exact Hugging Face versions are provided for reproducibility. We filter out samples without textual or visual prompts, as such samples are incompatible with our proposed semantic-equivalent perturbation and are therefore excluded. Relevant implementation details are summarized in Table~\ref{tab:supp_benchmark_detail}. For multiple-choice questions, we standardize all choice markers to numbers, \emph{e.g.}, 0, 1, ..., to simplify the subsequent answer-checking process. For MM-Vet, we include an additional prompt to prevent the LVLM from generating excessive irrelevant details, as the ground-truth answers in MM-Vet are relatively concise.

\noindent \textbf{Utilized LLM details.} We utilize Qwen2.5 as the LLM in our experiments (see Table~\ref{tab:supp_llm}). The LLM sizes range from 0.5B to 7B. The specific Hugging Face versions are provided. Qwen2.5 inherently supports multi-GPU inference. The LLM is employed for text-related operations during experiments, such as answer correctness checking, semantic entailment checking, and textual perturbation. The hyper-parameter settings are provided in Table~\ref{tab:supp_llm_hp}. A low temperature is used for answer correctness checking and semantic clustering, while a high temperature is applied during textual prompt perturbation.

\section{Hallucination Detection Cost Analysis}

We provide statistics on the hallucination detection memory and time costs for the adopted LVLMs (see Table.~\ref{tab:supp_cost}). GPU memory usage tends to increase with model size. Differences in model structure also affect GPU memory usage, \emph{e.g.}, LLaVA1.5 and LLaVANeXT exhibit different memory requirements despite having the same model sizes. For the time cost per sample in VL-Uncertainty hallucination detection, the overhead is low, especially considering the risks it can pinpoint and help avoid.

